\RequirePackage{fix-cm}
\documentclass{svjour3}                     
\smartqed  %
\usepackage{marginnote}

\usepackage{amssymb}
\usepackage{xcolor}
\usepackage{color}

\usepackage{rotating}
\usepackage{lscape}
\usepackage{url}

\usepackage{algorithmic}
\usepackage[ruled,lined,linesnumbered,longend]{algorithm2e}
\usepackage{float}
\usepackage{subfig}
\usepackage{amsmath}

\usepackage{natbib}

\journalname{Memetic Computing}

\begin{document}

\title{Deep Memetic Models for Combinatorial Optimization Problems: Application to the Tool Switching Problem}
\titlerunning{Deep Memetic Models for COPs: Application to the ToSP}

\author{Jhon Edgar Amaya         \and
        Carlos Cotta \and
        Antonio J. Fern\'andez-Leiva \and
        Pablo Garc\'ia-S\'anchez
}
\institute{
    Jhon Edgar Amaya \at 
    Universidad Nacional Experimental del T\'achira (UNET), Laboratorio de Computaci\'on de Alto Rendimiento (LCAR), San Crist\'obal, Venezuela
    \email{jedgar@unet.edu.ve}
    \and
    Carlos Cotta, Antonio J. Fern\'andez-Leiva \at
    Dept. Lenguajes y Ciencias de la Computaci\'on, ETSI Inform\'atica, University of Malaga, Campus de Teatinos, 29071 - Malaga, Spain
    \email{ccottap@lcc.uma.es, afdez@lcc.uma.es} 
    \and
    Pablo Garc\'ia-S\'anchez \at
    Dept. de Ingenier\'ia Inform\'atica, ESI, University of C\'adiz, Campus de Puerto Real, 11519 - C\'adiz, Spain
    \email{pablo.garciasanchez@uca.es}
}

\date{Received: date / Accepted: date}
\maketitle

\begin{abstract}
Memetic algorithms are techniques that orchestrate the interplay between population-based and trajectory-based algorithmic components. In particular, some memetic models can be regarded under this broad interpretation as a group of autonomous basic optimization algorithms that interact among them in a cooperative way in order to deal with a specific optimization problem, aiming to obtain better results than the algorithms that constitute it separately. Going one step beyond this traditional view of cooperative optimization algorithms, this work 
tackles deep meta-cooperation, namely the use of cooperative optimization algorithms in which some components can in turn be cooperative methods themselves, thus exhibiting a deep algorithmic architecture.
The objective of this paper is to demonstrate that such models can be considered as an efficient alternative to other traditional forms of cooperative algorithms.
To validate this claim, different structural parameters, such as the communication topology between the agents, or the parameter that influences the depth of the cooperative effort (the depth of meta-cooperation), have been analyzed. To do this, a comparison with the state-of-the-art cooperative methods to solve a specific combinatorial problem, the Tool Switching Problem, has been performed.
Results show that deep models are effective to solve this problem, outperforming metaheuristics proposed in the literature.

\end{abstract}

\section{Introduction \label{sec:introduction} }

\begin{sloppypar}
The optimization of combinatorial optimization problems (COPs) has been addressed from different algorithmic approaches. Initially, exact/complete search methods
showed good results to cope with  problem instances of  limited size, but they are also inefficient when the size of the problem scales up. Subsequently, other approaches called {\em metaheuristics}, 
such as bio-inspired optimization techniques, have shown a good performance to obtain high quality solutions at the expense of not proving optimality.
\end{sloppypar}

One of the key factors of the success of metaheuristics methods to deal with complex COPs is the balance between exploitation and exploration of the search space. %
However, finding the best balance between these two facets of the search is in fact another optimization problem, and not an easy one to solve.
An interesting proposal found in population-based optimization methods to tackle this issue is the use of structured populations, thus limiting interactions between individuals and allowing a better exploration and exploitation of the search space \citep{Lim2014}.

{\color{black} On the other hand}, {\em hybridization} %
basically means a synergistic union of different components, each one of them contributing different features to the search process and/or providing mechanisms for the exploitation of problem knowledge. From a broad perspective, hybridization can include the use of any problem-specific add-ons providing problem-knowledge (e.g., specialized decoders, ad-hoc variation operators, etc.), thus enhancing the search process. It is often the case that hybridization is used with a more specific connotation though, namely the combination of higher-level algorithmic components (frequently, techniques that could be used as stand-alone methods, such as other metaheuristics).
Under this latter prism, hybridization can be viewed from two wide perspectives \citep{DBLP:journals/eor/JourdanBT09}: integration and cooperation. %
Integration refers to the addition of one of the optimization techniques as a component of another optimization method
whereas cooperation generally relates to setting up a mechanism to exchange information between methods that are applied one after another or in parallel. The interest for this approach to optimization dates back to the 1990s
and, indeed, it can be considered itself a programming paradigm featuring two main elements \citep{Crainic2008}:
(a) a set of autonomous programs, each implementing a particular solution method, and (b) a cooperative scheme that combines these autonomous elements in a simple and unified strategy for optimization.

The cooperative optimization approach described above thus amounts to the application of various algorithmic components,
each one exploring a specific search landscape through processes of intensification/diversification ---inherent to the metaheuristics used--- to obtain an effective mechanism with the ability of escaping from local minima by exchanging information about the search space being explored. %
These collaborative optimization models constitute a very appropriate framework for integrating different search techniques: each one may exploit problem knowledge in a complementary way, and have a different view of the search landscape. Therefore, if we combine their different exploration patterns, the search benefits from new ways to avoid local optima. In fact, this feature is more useful whenever the problem addressed raises a challenging optimization task to each one of the individual search algorithms since otherwise computational power might be diversified spending time in unproductive or duplicated explorations.

Building on previous work on this kind of models \citep{Amaya2011MC} -- see also Section \ref{sec:backgrCoop} -- this paper tackles deep meta-cooperation, i.e., cooperative models in which at least one of its components is a cooperative model itself. {\color{black}Therefore, we bring together the notion of cooperative optimization sketched before with the idea of deep metaheuristics \citep{DBLP:journals/fgcs/CamachoLGCCLVO18} resulting in an unified model for building powerful algorithmic complexes for optimization.} {\color{black} While admittedly interpretable from different angles,} meta-cooperation naturally fits the broad memetic paradigm, since it constitutes a framework for arranging the interoperation among different methods, either based on populations or in local search, aiming to have synergistic effects by virtue of the adequately diversified exploitation/exploration capabilities of the techniques involved.
In this line, this work highlights this idea by proposing in Section \ref{sec:cooperativeModel} a basic schema that can be easily instantiated into a number of different algorithms, and provides evidence that deep meta-cooperative algorithms are effective optimization methods to deal with combinatorial problems, capable of outperforming shallow cooperative methods. {\color{black} Being a heterogeneous hierarchical agent-based system, this model exhibits several advantages. Firstly, it eases escaping from local optima by the combination of different search patterns. Secondly, it can be parallelized in a trivial way, following a island-based approach. Thirdly, it naturally lends itself to incorporating problem-knowledge in a flexible way, via any of the algorithmic components involved in the system. Thus, the main contributions of this work are (i) the proposal and definition of a new model for {\em deep meta-cooperation}, (ii) its application to a hard combinatorial problem (the ToSP, see below), outperforming state-of-the-art methods for said problem, and (iii) a performance study involving some salient design factors.}

To validate {\color{black} the proposed model}, the performance of a number of different algorithms (devised from the basic scheme,  with varied depths of meta-cooperation) have been analyzed in Section \ref{sec:experiments}, using a hard problem from the area of flexible manufacturing as test bench. The experimental analysis conducted considers issues such as the depth of the architecture and its spatial arrangement. We close the paper with a discussion of the main conclusions and an outline of future work in Section. \ref{sec:conclusions}.

\section{Background \label{sec:backgrCoop}}

This section is devoted to discuss some concepts related to cooperative models, as well as to provide an overview of the problem addressed in the experimental section, namely the Tool Switching Problem (ToSP), and how it has been approached in the literature.

\subsection{Cooperative Models}

In general, cooperative algorithms can be classified from different points of view; for instance, \cite{ElAbd2005} proposed a division based on two approaches: by considering the diversity of the underlying algorithms in the cooperative model or by the level of cooperation among them. The first approach focuses on the nature of algorithms according to their implementations and  catalogues them in two categories: (a) homogeneous,  cooperative algorithms that only employs one type of algorithm in all its agents, vs. (b) heterogeneous, where different algorithms are used. 

In addition, and independently of its homogeneous/heterogeneous nature, the underlying algorithms in a cooperative method can be executed in parallel or sequentially.
The second approach is based on the decomposition of the search space by some kind of mechanism. Three categories are proposed to classify the algorithms: {\em explicit} (i.e., the underlying algorithms %
explore different parts of the search space and interchange information via some communication strategy), {\em implicit} (i.e., the underlying algorithms  explore parts of the search space that were explicitly determined by some decomposition policy) or {\em hybrid} (i.e., a mixture of the previous ones). 

Also, from another point of view, two main models to implement cooperation schemes \citep{Crainic2008} have been discussed: (a) the {\em collaborative approach}, in which cooperation consists of a set of autonomous programs, usually called {\em agents}, that contain particular solution methods and that communicate among them to share information, and (b) the {\em integrative approach} that basically combines these autonomous elements in a simple and unified strategy to handle a problem. Each agent, in theory, should be able to address the problem independently. Next, we describe some the works that have tackled both approaches.

\subsubsection{Collaborative approaches}
One of the first examples of collaborative algorithms is the ``go with the winners'' approach by \cite{aldous1994go}, in which successive synchronization epochs are used to focus the search of a collection of otherwise independent trajectory-based algorithms. 

\begin{sloppypar}
\cite{Gallardo2007} proposed the intertwined use of a truncated branch-and-bound algorithm (akin to beam search) with evolutionary and memetic algorithms: the former would identify promising parts of the search space and would pass the information to the latter, that would use it to initialize their population and explore such regions. \cite{Talbi2006} proposed COSEARCH, in which three metaheuristics were used: tabu search (TS), genetic algorithm (GA) and kick operator (KO). The TS was used as control algorithm, the GA was focused on the diversification process and the KO was applied to intensify the search. The cooperative mechanism includes an adaptive memory to store and distribute the solutions. 

More aligned with the agent-perspective mentioned before,
\cite{Malek2009} introduced a multi-agent system.
Here the cooperative process is achieved by combining individual metaheuristics, implemented by individual agents, and exchanging solutions between them. Based on this idea,  \cite{Amaya2011MC} described a cooperative model that can be instantiated by different communication topologies to share solution candidates, and where each agent in the cooperative model can be loaded with a particular search algorithm. {\color{black} Another model based on autonomous agents was described by \cite{Byrsky13asymptotic}. In this work, individuals are active agents of the population, and not passive data structures (as in a canonical EA), characterized by a genotype and a meme that may change during the evolutionary process. Different types of actions were tested, including a mathematical formulation to analyze the interactions between agents.}%
\end{sloppypar}

Newer approaches, such as Multi-Agent Collaborative Search (MACS) \citep{Vasile2017} are also focusing on multi-objetive optimization, based on moving the agents to different dominant regions of the search space, and using weights for each objective function. Also, \cite{Fernandez-Leiva2018} have recently proposed an  algorithmic model to distribute a number of (interactive/proactive) User-centric Memetic Algorithms (UcMAs) that act as independent agents and, eventually, synchronize to interchange information. An experimental study shown that some distributed UcMAs, that were generated from the proposed model, especially those based on proactivity, performed efficiently in some hard combinatorial problems.

\subsubsection{Integrative approaches}
In the case of integrative approaches, different methods are combined into a unified strategy. {\color{black} For example, among the first papers on this subject, \cite{Anandalingam92hierarchical} propose the use of multi-agent systems with different objective functions that work in a non-cooperative way and follow the same restrictions, but there is a certain type of hierarchy when it comes to influencing other agents.} The inclusion of local-search components in population-based techniques typical of traditional memetic algorithms is {\color{black}another} immediate example, as shown in next subsection. 

From a different perspective,
\cite{Crainic2004} used a method with multiple searches as cooperative strategy. In particular, variable neighborhood search (VNS) was employed as a local search (LS) method with a central memory mechanism that managed the information exchange; different VNS instances cooperated asynchronously through the exchange of information to solve different instances of the p-median problem.
\cite{Cruz2009} proposed a cooperative strategy based on several problem solving strategies, called \emph{solvers}, and a supervisor, named \emph{coordinator}; the coordinator verifies when the solver provides new information and decides if the behavior of the solver must be adapted using a \emph{rule base}. They reported good results when this scheme was also applied to the p-median problem. 

\subsubsection{The Memetic Perspective: Toward Deep Architectures}
\begin{sloppypar} One issue that is crucial in cooperative algorithms is the adequate management of the balance between exploration and exploration of the search space. An efficient cooperative algorithm must deal with it; in this sense, this is one of the central tenets of memetic algorithms (MAs) \citep{Neri2012d}. %
In the so-called narrow vision of MAs, these techniques can be regarded as a combination of population-based optimization algorithms (that provide the exploration/diversification mechanism) and some form of local search (that is used for the exploitation/intensification of the search in certain parts of the search space). This can be readily generalized to a broader definition of MAs as methods that orchestrate the synergistic interplay of algorithmic components of global search and local search \citep{Neri2012d}.
Note that this idea also lies at the core of the notion of \emph{memetic computing}, which revolves about the harmonic 
coordination of complex computational structures composed of interacting modules (i.e., memes) for problem solving.
Memetic computing thus places more emphasis on the explicit management of memes, something which was already anticipated in the early stages of the paradigm and put to work in the early 2000s \citep{krasnogor2001emergence}. 
\end{sloppypar}

This idea of orchestrating an adequate combination of algorithmic components, central to MAs and related techniques as described before, gives rise naturally to the notion of \emph{deep metaheuristics}, cf.  \citep{DBLP:journals/fgcs/CamachoLGCCLVO18}: analogously to  deep learning algorithms (\cite{LBH15}; {\color{black} \cite{Cui18malicious}}), which feature multiple processing layers to learn representations of data with multiple levels of abstraction, deep bioinspired algorithms (and deep metaheuristics in general) exhibit multiple interconnected layers (or components, since generally speaking there needs not be a well-defined layered hierarchy {\color{black} \citep{Cui19cnns}}) which contribute different desired features to the search process by encapsulating the tools required to tackle the different aspects of the complexity of the problem (and even those of the computational substrate), and whose interaction optimize the solving process. There are different reasons why metaheuristics in general (and population-based techniques in particular) are amenable to deep architectures {\color{black} \citep{Cui19cyberphisical}}. They are flexible search paradigms whose plasticity allows accommodating multiple components, and whose inherent resilience and decentralization (a particularly relevant property of population-based algorithms) favors their useful integration.

\subsection{The Tool Switching Problem}

The Tool Switching Problem (ToSP) is a hard optimization problem present in many flexible manufacturing systems. It considers a reconfigurable machine that has to process a number of jobs, each of them requiring a certain collection of tools for being accomplished. To this end, the machine is endowed with a magazine composed of a number of slots into which different tools can be loaded (each slot supports just one tool). Jobs are executed sequentially, and therefore each time a job is being processed, the corresponding tools must be loaded on the magazine, whose space is limited. The crux of the problem difficulty is the fact that the total number of the tools for processing all jobs is greater than the number of slots available in the magazine, and therefore, all of them cannot be simultaneously present in the magazine. This means that it is eventually
necessary to perform a tool change, that is, some tool must be removed from the magazine and other tool has to replace it. In this context, the management of the tool changes is a challenging task that directly affects the efficiency of flexible manufacturing systems: an inadequate planning of jobs and/or a poor tool changing policy can result in excessive delays in the reconfiguration of the magazine as well as in the termination of the jobs. %

Although the order of the tools in the magazine is normally not relevant, the need for a tool switching, in fact, depends on the order in which the jobs are run. That is, the objective of the ToSP consists of finding a suitable job sequence in which jobs will be executed, and the corresponding sequence for loading/unloading the tools so that the number of tool switchings is minimized (this rests on the assumption that the time needed to change a tool is a significant part of the processing time of all jobs  and therefore, the tool changing policy significantly affects system performance).

{\color{black}
\subsubsection{Formal Representation} \label{sub:mathToSP}

Two major elements are present in the definition of the ToSP: a machine $M$ and a collection of jobs $J=\{J_{1},\cdots,J_{n}\}$
to be processed. The optimization process is guided by the tools required
for each job. Also, there is a set of tools $T=\{\tau_{1},\cdots,\tau_{m}\}$,
and that each job $J_{i}$ requires a certain subset $T^{(J_{i})}\subseteq T$ of
tools to be processed. With regard to the machine, we will just consider the capacity $C$ of the magazine, that is, the
number of available slots. From the previous elements, 
the ToSP can be formalized as follows: let a ToSP instance be represented by a pair,
$I=\langle C,A\rangle$ where $C$ denotes the capacity of magazine,
and $A$ is a $m\times n$ binary matrix that defines the tool requirements
to execute each job, i.e., $A_{ij}=1$, if and only if, tool $\tau_{i}$
is required to execute job $J_{j}$, being $0$ otherwise \footnote{We assume that $C<m$, otherwise the problem is trivial.}.

The solution
to such an instance is a sequence $\langle J_{i_{1}},\cdots,J_{i_{n}}\rangle$
, where $i_{1},\ldots,i_{n}$ is a permutation of numbers $1,\ldots,n$,
determining the order in which the jobs are executed, and also a sequence
$T_{1},\cdots,T_{n}$ of tool configurations ($T_{i}\subset T$) determining
which tools are loaded in the magazine at a certain time. Note that
for this sequence of tool configurations to be feasible, it must hold
that $T^{(J_{i_{j}})}\subseteq T_{j}$.

Let $\mathbb{N}_{h}=\{1,\cdots,h\}$ henceforth. We will index jobs and tools
 with integers from $\mathbb{N}_{n}$ and $\mathbb{N}_{m}$ respectively.
An integer linear programming (ILP) formulation for the ToSP is shown
below, using two sets of zero-one decision variables:
\begin{itemize}
\item $x_{jk}=1$ if the job $j\in\mathbb{N}_{n}$ is assigned to position $k\in\mathbb{N}_{n}$
in the sequence, and 0 otherwise ---see Eqs. (\ref{eq:bard2}) and
(\ref{eq:bard3})---,
\item $y_{ik}=1$ if the tool $i\in\mathbb{N}_{m}$ is in the magazine at instant
$k\in\mathbb{N}_{n}$, and 0 otherwise ---see Eq. (\ref{eq:bard4})---.
\end{itemize}

Processing each job requires a particular collection of tools loaded
in the magazine. Also, we assume that no job requires a number of tools
higher than the magazine capacity, i.e., $\sum_{i=1}^{m}A_{ij}\leqslant C$
for all $j\in\mathbb{N}_{n}$. Tool requirements are reflected in
(\ref{eq:bard5}). Following the work by \cite{Bard1988}, we
assume the initial condition $y_{i0}=1$ for all $i\in\mathbb{N}_{m}$.
This initial condition amounts to the fact that the initial loading
of the magazine is not considered as part of the cost of the solution.
The objective function $F(\cdot)$ counts the number of tool changes needed to complete a particular job sequence ---see Eq. (\ref{eq:bard1})---.
We assume that the cost of each tool switching is unitary and
constant.\medskip{}

\begin{equation}
\min\ F(y)=\sum_{j=1}^{n}\sum_{i=1}^{m}y_{ij}(1-y_{i,j-1})\label{eq:bard1}
\end{equation}

\begin{equation}
\forall j\in\mathbb{N}_{n}:\ \sum_{k=1}^{n}x_{jk}=1\label{eq:bard2}
\end{equation}

\begin{equation}
\forall k\in\mathbb{N}_{n}:\ \sum_{j=1}^{n}x_{jk}=1\label{eq:bard3}
\end{equation}

\begin{equation}
\forall k\in\mathbb{N}_{n}:\ \sum_{i=1}^{m}y_{ik}\leqslant C\label{eq:bard4}
\end{equation}

\begin{equation}
\forall j,k\in\mathbb{N}_{n}\ \forall i\in\mathbb{N}_{m}:\ A_{ij}x_{jk}\leqslant y_{ik}\label{eq:bard5}
\end{equation}

\begin{equation}
\forall j,k\in\mathbb{N}_{n}\ \forall i\in\mathbb{N}_{m}:\ x_{jk},y_{ij}\in\{0,1\}\label{eq:bard6}
\end{equation}

\medskip{}

This general definition shown above corresponds to the uniform ToSP,
in which each tool is adjusted to a single slot and all slots are
considered the same size. The relevant decision is whether the tool
is to be loaded or not in the magazine at any given time. %
Assuming this, if the job sequence is fixed,
the optimal tool switching policy can be determined in polynomial
time using a greedy procedure termed%
\emph{ Keep Tool Needed Soonest} (KTNS) \citep{Bard1988,Tang1988}.
The functioning of this procedure is as follows:
\begin{itemize}
\item At any instant, insert all the tools that are required for the current
job.
\item If one or more tools are inserted and there are no vacant slots on
the magazine, keep the tools that are needed soonest. Let $J=\langle J_{i_{1}},\cdots,J_{i_{n}}\rangle$
be the job sequence, and let $T_{k}\subset\mathbb{N}_{m}$ be the
tool configuration at time $k$. Let $\Xi_{jk}(J)$ be defined as
\begin{equation}
\Xi_{jk}(J)=\min\left\{ t\ |\ (t>k)\wedge\left(A_{jJ_{i_{t}}}=1\right)\right\} ,
\end{equation}

that is, the next instant after time $k$ at which tool $\tau_{j}$
will be needed again given sequence $J$. If a tool has to be removed,
the tool $\tau_{j^{*}}$ maximizing $\Xi_{jk}(J)$, is chosen, i.e.,
remove the tools whose next usage is furthest away in time.

\end{itemize}

The importance of this policy is that, as mentioned before, given
a job sequence KTNS obtains its optimal number of tool switches.

\subsubsection{ Neighborhood Function} \label{sub:NieghToSP}

The permutational representation of the problem allows to define different
structures, in particular, we focus in this paper on two structures, namely:
\begin{enumerate}
\item The well-known swap neighborhood ${\cal N}_{swap}(\cdot)$, in which
two permutations are neighbors if they just differ in two positions
of the sequence, that is, for a permutation $\pi\in\mathbb{P}_{n}$.

\begin{equation}
{\cal N}_{swap}(\pi)=\{\pi'\in\mathbb{P}_{n}\ |\ H(\pi,\pi')=2\}
\end{equation}

where $H(\pi,\pi')=n-\sum_{i=1}^{n}[\pi_{i}=\pi'_{i}]$ is the Hamming
distance between sequences $\pi$ and $\pi'$ (the number of positions
in which the sequences differ), and $[\cdot]$ is Iverson parenthesis
(i.e., $[P]=1$, if $P$ is true, and $[P]=0$ otherwise). Given the
permutational nature of sequences, this implies that the contents
of the two differing positions have been swapped.

\item The block neighborhood ${\cal N}_{block}(\cdot)$, a generalization
of the swap neighborhood in which a permutation $\pi'$ is a neighbor
of permutation $\pi$, if the former can be obtained from the latter
via a random block swap. A random block swap is performed as follows:

\begin{enumerate}
\item A block length $b_{l}\in\mathbb{N}_{n/2}$ is uniformly selected at
random.
\item The starting point of the block $b_{s}\in\mathbb{N}_{n-2b_{l}}$ is
subsequently selected at random.
\item Finally, an insertion point $b_{i}$ is selected, such that $b_{s}+b_{l}\leqslant b_{i}\leqslant n-b_{l}$,
and the segments $\langle\pi_{b_{s}},\cdots,\pi_{b_{s}+b_{l}-1}\rangle$
and $\langle\pi_{b_{i}},\cdots,\pi_{b_{i}+b_{l}-1}\rangle$ are swapped.
\end{enumerate}

Obviously, if the block length $b_{l}=1$ then the operation reduces
to a simple position swap, but this is not typically the case.

\end{enumerate}

\subsubsection{Tackling the ToSP} 
}
Different examples of this problem can be found in diverse fields, such as electronic products manufacturing, metallurgy, and aeronautics, among others  %
\citep{Bard1988,Tang1988}.
Note that the ToSP is an extremely difficult problem. Indeed, the scale of difficulty depends on the number of jobs, the number of tools and the capacity of the magazine. This problem has been addressed with exact methods 
but with moderate success, as the ToSP has been shown as NP-hard problem when the capacity of the magazine is greater than two (as is the usual case). Hence, exact methods are inherently limited to tackle this problem

The ToSP has been addressed by considering different
problem perspectives, such as an Integer Linear Programming (ILP) formulation \citep{Tang1988}, a non-linear integer program with a dual-based relaxation heuristic  \citep{Bard1988}, and
a number of other methods such as exact algorithms \citep{Laporte2004}, 
and (non-cooperative) meta-heuristics  \citep{AlFawzan2003,Amaya2011edam,Amaya2011ijc}.

Focusing on cooperative methods, \cite{Amaya2010n} proposed a hybrid cooperative model that used spatially-structured agents using specific local-search and population-based strategies, showing that this
 model provided better results than their constituent parts. Later,
\cite{Amaya2011MC} used a cooperative architecture where several agents cooperated in solving the same problem by using genetic algorithms in conjunction with both tabu search and hill climbing. This proposal benefited from maintaining a high diversity of possible solutions, as well as from providing a certain degree of independence in the exploration of different regions of the search space, similarly to the case of island-based evolutionary systems. Some instances of the proposed model performed significantly better than state- of-the-art methods. Going beyond these results, next section describes a generic meta-cooperative proposal which will be subsequently deployed on the ToSP.\\

\section{A Framework for Deep Meta-Cooperation \label{sec:cooperativeModel}}

The model proposed consists of the hierarchical aggregation at several levels of a pool of algorithmic agents --loaded with a particular metaheuristic and random initial solutions-- that concurrently explore different parts of the search space. The search process is performed by each agent for a particular amount of time (or number of evaluations), after which the best solution found by each agent is distributed by the specific cooperative communication topology model at each level in which nested cooperation takes place. To formalize this idea, let us describe a basic template for cooperation which can then be recursively expanded into a certain deep architecture.

The basic cooperation template can be described following different taxonomies or models presented in the literature. Thus, it can be framed in the classification proposed by \cite{ElAbd2005} as a serial-heterogeneous scheme with implicit decomposition of the search space. It can also align with the DACOS system proposed by \cite{DBLP:journals/spe/AmoPMV10} for the configuration and analysis of centralized cooperative systems. 
Let $\mathbb{N}_{n}=\{1,\cdots,n\}$, and let $R$ be an architecture with $n$ agents, where each agent $a_{i}$ ($i\in\mathbb{N}_n$) in $R$ contains a metaheuristic, which can be a local search method, a population-based method, a hybrid technique or any other cooperative scheme.
These agents engage in isolated exploration periods followed by synchronous communication.
We denote as $cycles_{\max}$ the maximum number of cycles of exploration/communication in this cooperative model.

\begin{algorithm}[!t]
\floatname{algorithm}{Algoritmo} %
\caption{Cooperative model} \label{code:Cooperative} %
\setcounter{AlgoLine}{0}
\tcp{Generate the initial solutions for each component}
\For{$i\in\mathbb{N}_{n}$} { \tcp{In the case of LS method only one solution is generated (i.e. $\#S_i =1$). In population-based algorithms, $\#S_i > 1$} $S_i \leftarrow$ \textsc{GenerateInitialPopulation()}\; } 

$E_{cycle} \leftarrow E_{\max}/cycles_{\max}$

$cycles \leftarrow 1$\; 
\BlankLine 
\While{$cycles \leqslant cycles_{\max}$} { \For{$i\in\mathbb{N}_{n}$} { $S_i \leftarrow {a}_i(E_{cycle}/n)$; \tcp{Apply the method in ${a}_i(.)$ during $E_{cycle}/n$ evaluations and return the solution/s found.} } \For{$(i,j)\in \Lambda$} {
\tcp{Minimization assumed w.l.o.g.}
\If{$fitness($\textsc{Best}$(S_i)) < fitness($\textsc{Best}$(S_j))$} { $S_j \leftarrow S_j \cup \{$\textsc{Best}$(S_i)\} \setminus \{$\textsc{Worst}$(S_j)\} $; \tcp{new accepted solution} } } $cycles \leftarrow cycles + 1$\; }
\Return{$\arg\min\{fitness($\textsc{Best}$(S_i)) \ |\ i \in \mathbb{N}_n\}$}\;

\end{algorithm}

Now, let $S_{i}$ be the set of candidate solutions generated by agent $a_{i}$ (i.e., by the execution of the metaheuristic\footnote{For example, if the agent is loaded with a local search method then only one solution will be generated and kept so that $\#S_{i}=1$, but if the agent is loaded with a population-based method ---for example, a classical memetic or genetic algorithm--- then a pool of candidate solutions will be generated so that $\#S_{i}\geqslant1$, where $\#S_{i}$ indicates the cardinality of $S_{i}$.} contained in the agent $i$), and let $\Lambda\subseteq\mathbb{N}_{n}\times\mathbb{N}_{n}$ be the communication topology\footnote{That is, if $(i,j)\in\Lambda$ then $a_{i}$ will send information to the agent $a_{j}$ in each cycle of synchronization inside the execution of the cooperative algorithm as described later in this paper.} in $R$.
The functioning of the model is described in Algorithm \ref{code:Cooperative}. First, all agents start with random initial solutions (lines 1-3). Then, the algorithm is executed over a maximum number of cycles (lines 6-16),
where in each cycle the search technique incorporated into each agent is executed to update its pool of associated solutions (lines 7-9). Of course, if the agent contains a LS method then only one solution will be in its solution pool and an improvement of this will be performed. However, if the agent has a population-based method then it generates a new set of solutions. Solutions are supplied from one agent to another, according to the topology considered (lines 10-14). An agent accepts a solution only if the received solution is better than the best solution stored by the agent in its corresponding set of candidate solutions (lines 11-13). Note that other policies to send and receive solutions (to and from the agents in the architecture) might be considered \citep{Nogueras2014Island}.

Regarding the communication topology, different settings can be used. Throughout this work the following ones have been considered:
\begin{itemize}
\item \textbf{Ring}. $\Lambda=\{(i,i(n)+1)\mid i\in\mathbb{N}_{n}\text{ and }i(n)\text{ denotes }i\text{ mod }n\}$. This imply a circular list of agents in which each node only sends information to its successor, and respectively, receives from its predecessor.
\item \textbf{Broadcast}. $\Lambda=\mathbb{N}_{n}\times\mathbb{N}_{n}$, i.e., a {\em go with the winners}-like topology.  Here, the best overall solution at each synchronization point is transmitted to all agents. This imply intensification over the same local region of the search space at the beginning of each cycle, performed by all the agents.
\item \textbf{Random}. $\Lambda$ is composed by $n$ pairs $(i,j)$ randomly sampled from $\mathbb{N}_{n}\times\mathbb{N}_{n}$. This sampling is performed each time the communication is carried out, and therefore, any of the two agents may eventually communicate at any synchronization point.
\end{itemize}

\begin{sloppypar}
The algorithm will be executed for a maximum number of evaluations $E_{\max}$ (value that has to be set initially) and for a given number of cycles $cycles_{\max}$ (also fixed initially). Note that  each cycle in our cooperative algorithm consumes $E_{cycle}=E_{\max}/cycles_{\max}$ evaluations. This computational effort is then shared (in a uniform way) among all the $n$ agents so that the specific search method of any agent consumes $E_{cycle}/n$ evaluations at most.
\end{sloppypar}

The template shown in Algorithm \ref{code:Cooperative} is thus parameterized by $n$, $E_{\max}$, $cycles_{\max}$, the set of metaheuristics loaded in agents $a_1,\ldots,a_n$, and the communication topology $\Lambda$. Hence, it is possible to generate many different actual instances of this template by adequately parameterizing the model.
In the following, $\Phi=cycles_{max}\Lambda(a_{1},a_{2},...,a_{n})$ will be used to denote a cooperative algorithm (an instance of the template shown in Algorithm~\ref{code:Cooperative})  composed by $n$ agents $(a_{1},a_{2},...,a_{n})$, a communication topology $\Lambda\in\{$\textsf{Ri(ng)}, \textsf{Br(oadcast)}, \textsf{Ra(ndom)}$\}$ and a maximum number of cycles $cycles_{max}$, and where each agent $a_i$ can be loaded with a specific optimization method.
For example, a Ring topology-based cooperative algorithm composed of three agents loaded with a steepest-ascent hill climbing method (\textsf{HC}), a simulated annealing technique (\textsf{SA})  and a tabu search algorithm (\textsf{TS}), and with a maximum number of 4 cycles is denoted as $\Phi=$ \textsf{4Ri(HC,SA,TS)}.%

As mentioned before, any agent might also be loaded with a cooperative algorithm as defined above, hence resulting in a  meta-cooperative algorithm. 
By extending the notation introduced previously, a meta-cooperative algorithm is represented as
$\Phi^{g}=cycles_{max}\Lambda(a_{1},a_{2},...,a_{n})$, where $g$ is the degree of meta-cooperation (or meta-cooperation depth). More precisely,
let ${\cal A}$ be a set of basic metaheuristics; the degree of meta-cooperation $g$ of a meta-cooperative model $\Phi^{g}=cycles_{max}\Lambda (a_{1},a_{2},...,a_{n})$ determines the depth of meta-cooperation, and is based on the following recursive definition:

\vbox{ %
\begin{itemize}
\item[(0)]  $\Phi^{0}=cycles_{max}\Lambda(a_1,\ldots,a_n)$ where 

\hspace{1cm}$\forall i\in\mathbb{N}_{n}: a_i \in {\cal A} $.

\item[\hspace{1cm}(1)] $\Phi^{1}=cycles_{max}\Lambda(a_1,\ldots,a_n)$ 
where 

\hspace{1cm}$\exists i\in\mathbb{N}_{n}:a_i=\Phi^{0}$ and 

\hspace{1cm}$\forall j\in\mathbb{N}_{n}: (a_j=\Phi^{0}$ or $a_j \in {\cal A})$

\item[$\vdots$]

\item[$(g)$] $\Phi^{g}=cycles_{max}\Lambda(a_1,\ldots,a_n)$ where 

\hspace{1cm}$\exists i\in\mathbb{N}_{n}:a_i=\Phi^{g-1}$ and 

\hspace{1cm}$\forall j\in\mathbb{N}_{n}: \exists g' < g: (a_j=\Phi^{g'}$ or $a_j \in {\cal A})$ 
\end{itemize}
}

\begin{figure}
\includegraphics[scale=0.5]{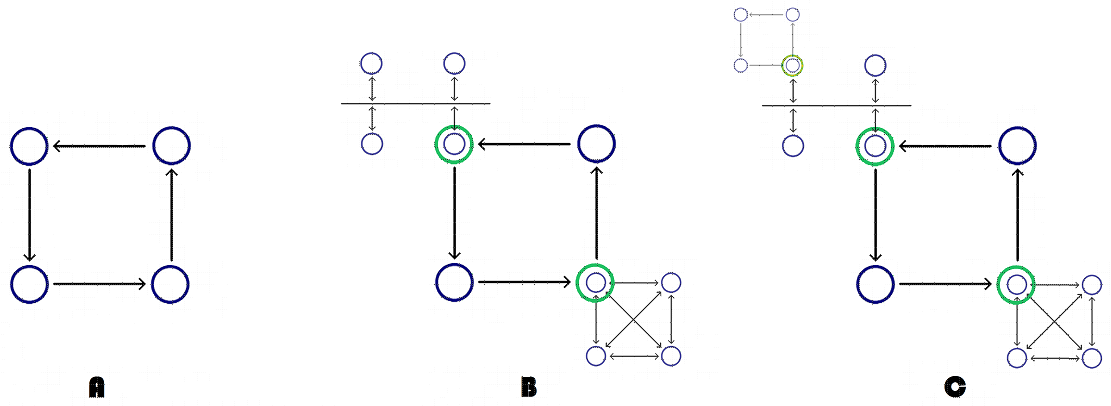}\caption{
\color{black}
Different instances of the meta-cooperative model: 0-level  ($\Phi^{0}$),  1-level ($\Phi^{1}$), and a 2-level ($\Phi^{2}$), respectively.
}
\label{fig:metalevels}
\end{figure}

As an example,
assume that ${\cal A}$ contains a classical steepest ascent hill climbing method (\textsf{HC}), a memetic algorithm with tabu search (\textsf{MATS}), and an iterated local search method ($ILS$).
Now, suppose we have three agents connected in a \textsf{Ring} topology, respectively loaded with \textsf{HC}, \textsf{MATS}  and a cooperative algorithm that in turn connects three agents in a \textsf{Broadcast} topology,
 all of them loaded with $ILS$. This latter cooperative algorithm is denoted as $\Phi^{0}=$ \textsf{Br(ILS,ILS,ILS)} and represents a $0$-level meta-cooperative algorithm, whereas the first mentioned algorithm is  denoted as $\Phi^{1}=$ \textsf{Ri(HC,MATS,$\Phi^{0}$)} and  represents a $1$-level meta-cooperative algorithm\footnote{For simplicity we have omitted the parameter $cycles_{max}$; this will be done in the following when necessary to improve the legibility.}.

  In general, $\Phi^{i}$ represents an $i$-level meta-cooperative algorithm or, in other words, an $n$-agent cooperative algorithm in which, at least, one agent contains a meta-cooperative algorithm of level $i-1$.
Therefore, the general architecture of a meta-cooperative algorithm simply corresponds to a particular instance of the cooperative model described in Algorithm \ref{code:Cooperative}. 
Figure \ref{fig:metalevels} shows different configuration examples of meta-cooperative algorithms. \sloppypar
{\color{black} Sub-figure \ref{fig:metalevels}(a) represents
a 0-level meta-cooperative model denoted as $\Phi^{0}=Ri(a_{1},a_{2},a_{3},a_{4})$
where $a_{i}$ is any simple metaheuristic (e.g., a population-based metaheuristic). Sub-figure \ref{fig:metalevels}(b) represents a 1-level meta-cooperative
model denoted as $\Phi^{1}=$ \textsf{Ri($a_{1},\Phi_{a}^{0},a_{3},\Phi_{b}^{0}$)} where both $\Phi_{a}^{0}=$ \textsf{Ra($a'_{1},a'_{2},a'_{3},a'_{4}$)}
and $\Phi_{b}^{0}=$ \textsf{Br($a''_{1},a''_{2},a''_{3},a''_{4}$)} represent 0-level meta-cooperative proposals.
Sub-figure \ref{fig:metalevels}(c) represents a 2-level meta-cooperative model denoted as $\Phi^{2}=$ \textsf{Ri($a_{1},\Phi_{a}^{0},a_{3},\Phi_{c}^{1}$)}
that contains a 0-level meta-cooperative algorithm ($\Phi_{a}^{0}$) in one agent and a 1-level meta-cooperative algorithm $\Phi_{c}^{1}=$ \textsf{Br($a''_{1},a''_{2},a''_{3},\Phi_{d}^{0}$)} loaded in other of its agents (and where $\Phi_{d}^{0}=$ \textsf{Ri($a'''_{1},a'''_{2},a'''_{3},a'''_{4}$)} is a 0-level meta-cooperative algorithm). 
}

\section{Experiments and Results}
\label{sec:experiments}

\subsection{Experimental setup \label{sec:expers}}

This section describes the experimental setup used for the evaluation of the meta-cooperative approach. For clarity, we firstly present considerations related to ToSP including benchmarks used. Later, the implemented algorithms involved in the experiments are discussed, as well as the considerations and parameters associated with each algorithmic proposal. Afterwards, specific scenarios will be implemented and put to test.

\subsubsection{Benchmarking \label{sec:bench}}

As there is no standard benchmark for ToSP, a wide range of problem instances considered in the literature \citep{Bard1988,Hertz1998,AlFawzan2003,Zhou2005} 
have been selected. More precisely, we chose 16 instances with number of jobs, machine capacity and number of tools in $[10,50]$, $[4,30]$ and $[9,60]$, respectively. The same instances were used in \citep{Amaya2010n,Amaya2011MC,Amaya2011edam,Amaya2011ijc}. Table \ref{tab:ToSPdatasets} shows the different problem instances chosen for the experimental evaluation.
Each instance is denoted as $C\zeta_{n}^{m}$ ($n$ jobs, $m$ tools and machine capacity $C$), and has different difficulty levels depending on the minimum and maximum number of tools required for all jobs.

\begin{table}[ht]
\caption{Instances of the problem considered in the experimental evaluation. The minimum and maximum number of tools required for all jobs, and the original work where they were published are indicated:
[A] \citep{AlFawzan2003}, [B] \citep{Bard1988}, [H] \citep{Hertz1998}, [Z] \citep{Zhou2005}.}
\begin{center}
\begin{small}
\begin{tabular}{ r  r  r  r  r  r  r  r  r }
\hline
        & $4\zeta_{10}^{9}$    &  $4\zeta_{10}^{10}$ & $6\zeta_{10}^{15}$  & $6\zeta_{15}^{12}$  &  $6\zeta_{15}^{20}$  &  $8\zeta_{20}^{15}$  & $8\zeta_{20}^{16}$  & $10\zeta_{20}^{20}$  \\
\hline
Min.   &  2 & 2 & 3 & 3 & 3  & 3 & 3 & 4\\
Max.   &  4 & 4 & 6 & 6 & 6  & 8 & 8 & 10\\
Source & [B,Z]      & [A,H] & [Z]  & [B,Z] & [H] & [A] & [B,Z] & [B,Z]\\
\\

\hline
       & $10\zeta_{30}^{25}$ & $15\zeta_{30}^{40}$ &  $15\zeta_{40}^{30}$ & $20\zeta_{40}^{60}$ & $24\zeta_{20}^{30}$ &  $24\zeta_{20}^{36}$ & $25\zeta_{50}^{40}$ & $30\zeta_{20}^{40}$  \\
\hline
Min.   & 4 & 6  & 6 & 7 & 9 & 9 & 9& 11 \\
Max.   &10 & 15 & 15& 20& 24& 24& 20& 30 \\
Source & [A] & [H] & [A] & [H] &  [B,Z] & [B,Z] & [A]& [Z] \\

\hline
\end{tabular}
\end{small}
\label{tab:ToSPdatasets}
\end{center}
\end{table}

Five different datasets%
\footnote{All the datasets are available at \url{http://www.unet.edu.ve/~jedgar/ToSP/ToSP.htm}} (i.e., relations among tools and jobs) were generated randomly per instance. The restriction that no job is covered by any other job, in the sense that no job (in the sequence of jobs to execute) is part of another different job, has been considered to avoid trivial solutions \citep{Bard1988,Zhou2005}.

\subsubsection{Metaheuristics \label{sec:meth}}

As previously explained, the meta-cooperative algorithms can be loaded with different  metaheuristics, either basic ones (those in ${\cal A}$) or other meta-cooperative algorithms. Regarding ${\cal A}$ we have considered the following methods, which have been used with success in the literature for solving the ToSP (we refer to the corresponding works for further details on these methods):

\begin{itemize}
\item {Hill Climbing algorithm (\textsf{HC})} \citep{Amaya2011edam}: In this algorithm,  the neighborhood 
is partially explored (only 4$n$ solutions, where $n$ is the number of jobs of the ToSP instance), and the best solution found is taken as the new current solution if it is better than the current one. If no best solution is found in the current neighbourhood, the search is considered stagnated, and can be restarted from a different initial point.
\item {Cross-entropy method with a single (\textsf{CE}) or with multiple (\textsf{CEM}) probabilistic mass functions} \citep{Amaya2011ijc}: In both methods $n^2$ solutions are generated by iteration cycle. The portion of elite solutions is defined as $\rho=0.01$. 
\item {Memetic evolutionary algorithm with tabu search (\textsf{MATS}) and with hill climbing (\textsf{MAHC})  } \citep{Amaya2011edam}: These two MAs are based on an elitist generational model.
The population size is 30, the crossover probability is set to 1.0 and mutation probability is defined as $1/\ell$, where $\ell$ is the length of the solution. A binary tournament selection is used, mutation uses random block swap, and recombination is based on alternating crossover position (APX)  -- recall solutions are permutation of jobs. The hybrid approach uses partial lamarckism, %
i.e., the local search is applied to a fraction of the individuals. The LS phase is applied to any individual with a probability $p_{LS}$; in case of application, the improvement uses up $LS_{evals}$ evaluations (or in the case of HC until stagnation, whatever occurs first). We used values of $LS_{evals}$ equals to 200 and $p_{LS}=0.01$.
\item {Basic Cooperative model $cycles_{max}\Lambda$\textsf{(MAHC,MATS,MAHC)}} \citep{Amaya2011MC}: This cooperative model has been defined in Section \ref{sec:cooperativeModel}. The memetic algorithms in the cooperative model uses the same parameters discussed before. Also, $cycles_{max}=5$ and $\Lambda=$ \textsf{Ring}.
\end{itemize}

We have implemented all metaheuristics and meta-cooperative models in Java. All experiments have been executed on a PC (Intel Celeron 1.5 GHz 512 MB) with Linux Debian (kernel 2.6.16-2-686). All algorithms were run $10$ times on each dataset. This means that each algorithm was executed 50 times per problem instance. Also, for a given problem instance 
$C\zeta_{n}^{m}$, a maximum number of evaluations per run $E_{max} = \varphi n(m - C)$ (with $\varphi > 0$) was set. That is, the maximum number of evaluations depends on the instance used, being $C$ the capacity of the magazine, $m$ the number of tools, and $n$ the number of jobs. Our previous experiments on the value of $\varphi$ proved that $\varphi = 100$ is an appropriate value that allows to keep an acceptable relation between computational cost and solution quality.

\subsubsection{Description of the scenarios}

In this section different comparative studies are performed regarding the basic parameters of the meta-cooperative model shown in Algorithm~\ref{code:Cooperative}. 
In each case, the results will be analyzed and compared with the best methods found until now. 

The experiments are designed under considerations proposed by \cite{Amaya2010n}: number of agents equals to $3$ and the maximum number of cycles is set to $5$. Additionally, only a limited set of basic algorithms (described in Section \ref{sec:meth}) are assessed. The main idea is to establish a framework to compare these algorithms in a fair way because of the combinatorial explosion of potential parameterizations. 

The experiments are presented in 3 scenarios aimed to determine which is the most adequate meta-cooperative algorithm for the ToSP. The experimentation is conducted in an incremental way: initially, we consider the best cooperative algorithm known until now (i.e., \textsf{5Ri(MAHC,MATS,MAHC)}) as a 0-level meta-cooperative algorithm, and use it as control algorithm (i.e., the technique with which to compare);  then, in the first scenario, we compare this control algorithm with the 1-level meta-cooperative algorithm providing the best performance in this scenario. The best of these two algorithms is considered as the control algorithm in the next scenario. %
We thus aim to analyze meta-cooperative algorithms with different meta-cooperation depth ($g \in \{1,2,3\}$).

\begin{sloppypar}
Note that the number of metaheuristics that can be loaded in the agents of a cooperative algorithm is finite in our experiments but there is a high number of possible combinations. %
So, to limit the number of experiments, only $5$ families of metaheuristics %
are considered, given the best results published until now. Thus, we will restrict our experimental analysis to the following algorithms: 
\begin{itemize}
    \item \textsf{5}$\Lambda$\textsf{(}$\Phi^{g-1}$\textsf{,CE,HC)},
    \item \textsf{5}$\Lambda$\textsf{(}$\Phi^{g-1}$\textsf{,CEM,HC)},
    \item \textsf{5}$\Lambda$\textsf{(}$\Phi^{g-1}$\textsf{,MAHC,CE)},
    \item \textsf{5}$\Lambda$\textsf{(}$\Phi^{g-1}$\textsf{,MAHC,CEM)} 
    \item \textsf{5}$\Lambda$\textsf{(}$\Phi^{g-1}$\textsf{,MAHC,HC)}
\end{itemize}
where $\Phi^{g-1}$ is a $(g-1)$-level-meta-cooperative model.
\end{sloppypar}

\begin{sloppypar}
These scenarios will guide us to determine the influence of the meta-cooperation depth in the ToSP
and whether there is some depth level in which noticeable degradation of performance ensues, that is, %
to understand when the performance decreases if we consider the next level of meta-cooperation. Moreover, a discussion comparing the different topologies used in the meta-cooperative models will be presented.
\end{sloppypar}

{\color{black} Table \ref{tab:parameters} summarize all the parameters used in the experiments.}

\begin{table}[!t]
    \centering
    {\color{black}\begin{tabular}{|l|l|}
    \hline
    \multicolumn{2}{|c|}{ \textbf{General}} \\ \hline
Capacity of the magazine	&	$C$	\\ \hline
Number of tools	&	$m$	\\ \hline
Number of jobs of the TSP instance	&	$n$	\\ \hline
$E_{max}$ 	&	 $\varphi n(m - C)$	\\ \hline
$\varphi$	&	100	\\ \hline
Runs per dataset	&	10	\\ \hline
Runs per instance	&	5	\\ \hline
Number of agents in each scenario	&	3	\\ \hline
$cycles_{\max}$	&	5	\\ \hline
\multicolumn{2}{|c|}{\textbf{Hill Climbing}}		\\ \hline
Solutions of the neirborhood	&	4n	\\ \hline
\multicolumn{2}{|c|}{\textbf{CE and CEM}}			\\ \hline
Solutions per iteration	&	$n^ 2$	\\ \hline
$\rho$	&	0.01	\\ \hline
\multicolumn{2}{|c|}{\textbf{MATS and MAHC}}			\\ \hline
Population size	&	30	\\ \hline
Crossover probability	&	1	\\ \hline
Mutation probability 	&	$1/\ell$	\\ \hline
$p_{LS}$	&	0.01	\\ \hline
$LS_{evals}$	&	200	\\ \hline
    \end{tabular}
    }%
    \caption{Summary of the parameters used in the experiments.}
    \label{tab:parameters}
\end{table}

\subsection{Results}
This section describes the obtained results of the performance of the algorithms in the different scenarios. To analyze the results a
rank-based approach has been used. This approach is based on computing the rank $r_j^i$ of
each algorithm $j$ on each instance $i$: rank 1 for the best, and
rank $k$ for the worst, where $k$ is the number of algorithms compared. That way we can compare the average rank of each algorithm, and perform appropriate non-parametric statistical tests.

\subsubsection{Basic Metaheuristics}

\begin{sloppypar}
Initially, we executed the basic metaheuristics discussed in Section \ref{sec:meth}. Each one of the five datasets was tested 10 times for each metaheuristic. The  results of executing the algorithms for the different instances can be seen in Table \ref{tab:resulBef}, showing the average results and standard deviation for each instance. This table also shows the different topologies used in the cooperative algorithm. It can be observed that the best algorithms are \textsf{CEM}, \textsf{CE}, \textsf{5Ri(MAHC,MATS,MAHC)} and \textsf{MAHC}. In particular, \textsf{CEM} is the one that obtains the best average results in all instances, consistently with the good behavior of this algorithm as reported in the literature. Note also that the primary goal of the experimentation with this scenario is obtaining the baseline performance of the algorithms, in order to dive into progressively larger architectural depth as shown in the following.  
\end{sloppypar}

\subsubsection{Scenario with 1-level-meta-cooperative model }

\begin{sloppypar}
The best cooperative model for the ToSP, i.e., $\Phi^{0}=$ \textsf{5Ri(MAHC,MATS,MAHC)} \citep{Amaya2010n}, has been selected in this scenario. For reasons of simplicity we will denote $\Phi^{0}$ from now on as \textsf{Hu}\footnote{{\sf Hu}, {\sf Ca} and {\sf Ox} are 1, 2 and 3 in Mayan language, respectively.}. Five metaheuristics combinations (or families) were generated for the  $1$-level-meta-cooperative model as aforementioned: \textsf{5}$\Lambda$\textsf{(Hu,CE,HC)}, \textsf{5}$\Lambda$\textsf{(Hu,CEM,HC)}, \textsf{5}$\Lambda$\textsf{(Hu,MAHC,CE)}, \textsf{5}$\Lambda$\textsf{(Hu,MAHC,CEM)} and \textsf{5}$\Lambda$\textsf{(Hu,MAHC,HC)}. In addition, each one of the five combinations was tested with each one of the three topologies considered in this paper (i.e., \textsf{Random}, \textsf{Ring} and \textsf{Broadcast}). Table \ref{tab:resulHu5} summarizes the results obtained by the set of 1-level-meta-cooperative algorithms (termed as $\Phi^{1}$) applied to optimize the 16 problem instances presented in Table~\ref{tab:ToSPdatasets}.
\end{sloppypar}

\begin{figure}[!ht]
\includegraphics[width=\textwidth]{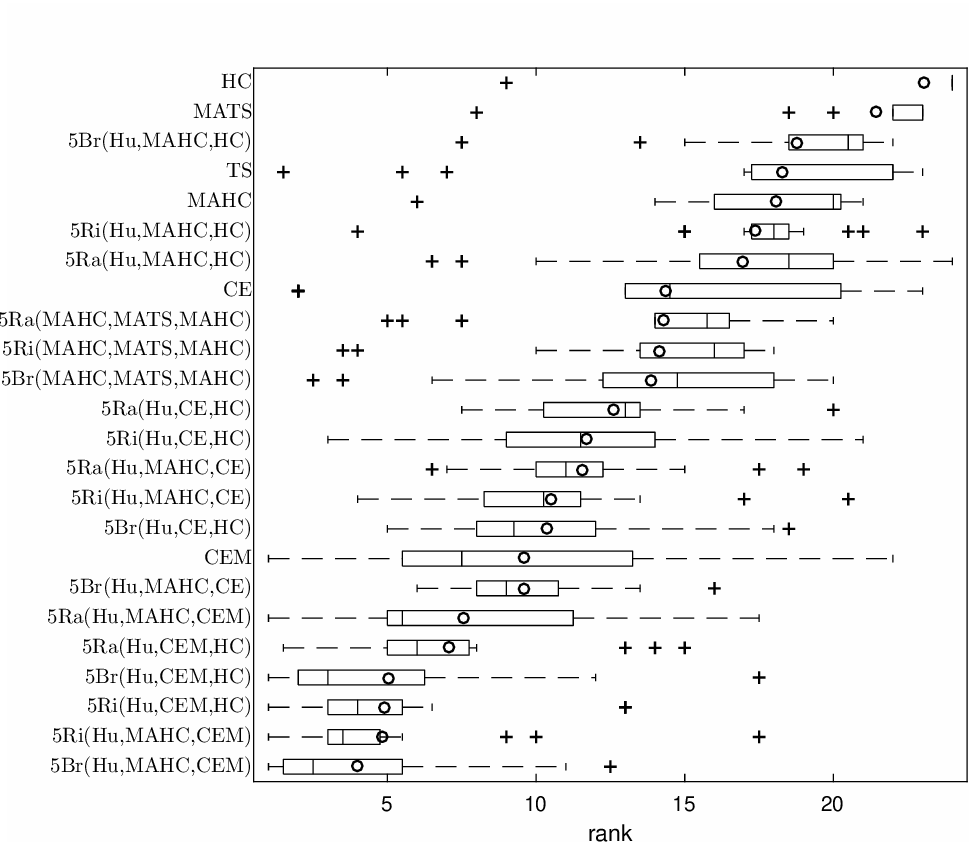}
\caption{Rank distribution for 1-level-meta-cooperative algorithms with respect to the best algorithmic approaches to solve the ToSP. In this figure and in subsequent ones, each box comprises the second and third quartiles of the distribution, the median (resp. mean) is marked with a vertical line (resp. circle), whiskers span 1.5 times the inter-quartile range, and outliers are indicated with a plus sign.
}
\label{fig:Fi1}
\end{figure}

\begin{sloppypar}
In general, it is observed that these meta-cooperative algorithms exhibit good performance, particularly in schemes \textsf{5}$\Lambda$\textsf{(Hu,CEM,HC)} and \textsf{5}$\Lambda$\textsf{(Hu,MAHC,CEM)}. To determine the behavior of the best meta-cooperative algorithms, a rank analysis was conducted, whose results are shown in Figure \ref{fig:Fi1}. Note how \textsf{5}$\Lambda$\textsf{(Hu,MAHC,CEM)} and \textsf{5}$\Lambda$\textsf{(Hu,CEM,HC)} obtained better results than \textsf{CEM}, the state-of-the-art algorithm in the literature and the best performing technique in the previous step.
\end{sloppypar}

Two well-known non-parametric statistical tests to compare ranks have been used: firstly, Quade test 
indicates that the statistical values obtained are considerably higher than the critical value, both when the whole collection of twenty-four algorithms (fifteen 1-level meta-cooperative approaches plus the nine basic algorithms considered in previous subsection) is considered, or just the top-ten performing ones ($p$-values $\approx 0$ in both cases). Therefore, the null hypothesis ---i.e. all algorithms are equivalent--- may be rejected with very high confidence. 
Subsequently, Holm test %
has been used to assess the relative performance of different algorithms
with respect to the control algorithm (the algorithm with the best average rank). The results are shown in Table \ref{statistical:holm} for the top-ten algorithms.
The control algorithm, i.e., \textsf{5Br(Hu,MAHC,CEM)} passes the test with a confidence level of $95\%$ for all algorithms except \textsf{5Ri(Hu,MAHC,CEM)}, \textsf{5Br(Hu,CEM,HC)} and \textsf{5Ri(Hu,CEM,HC)}. The difference with respect \textsf{CEM} is thus deemed significant. Note in any case the prevalence of \textsf{CEM} (a top-performing technique on itself) as a constituent part of the best cooperative models. This underpins their capability for enhancing in a synergistic way the performance of methods in lower levels of the cooperation hierarchy.

\begin{figure}[!ht]
\includegraphics[width=\textwidth]{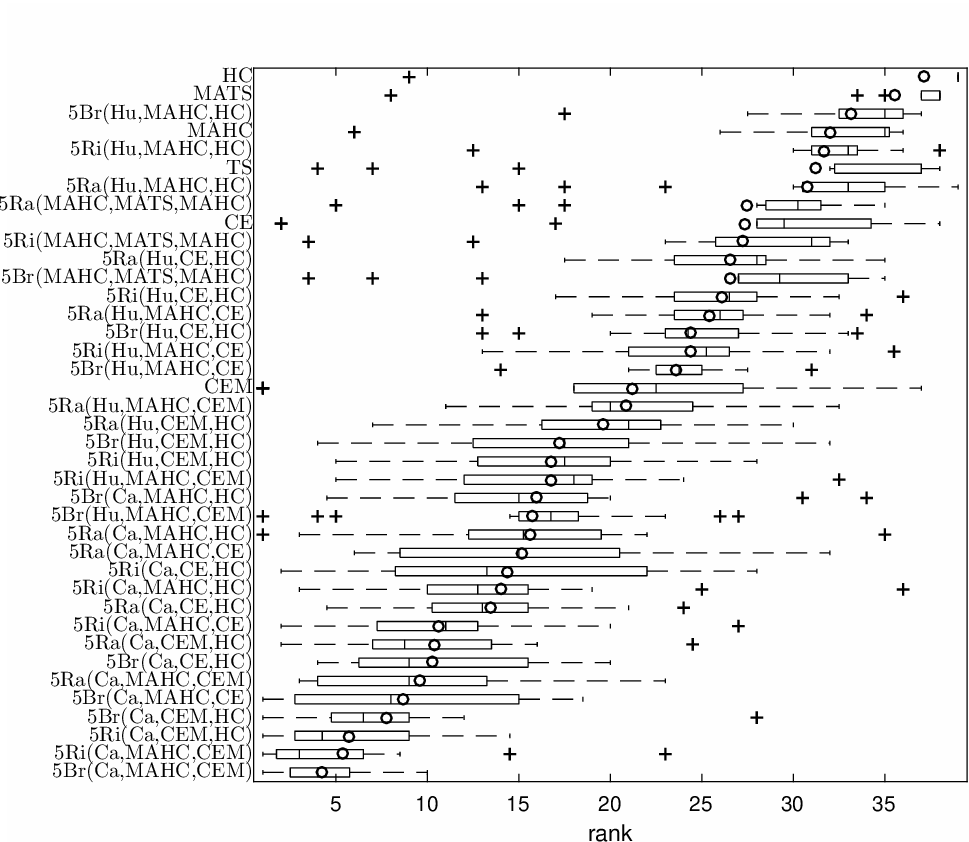}
\caption{Rank distribution for 2-level-meta-cooperative models. }
\label{fig:Fi1Fi2}
\end{figure}

\subsubsection{Scenario with 2-level-meta-cooperative models }

\begin{sloppypar}
Since 1-level-meta-cooperative results were better than those obtained by the best cooperative methods described so far, we proceed to perform an experimental analysis with the same 5 schemes \textsf{5}$\Lambda$\textsf{(}$\Phi^{1}$\textsf{,CE,HC)}, \textsf{5}$\Lambda$\textsf{(}$\Phi^{1}$\textsf{,CEM,HC)}, \textsf{5}$\Lambda$\textsf{(}$\Phi^{1}$\textsf{,MAHC,CE)},
\textsf{5}$\Lambda$\textsf{(}$\Phi^{1}$\textsf{,MAHC,CEM)} and \textsf{5}$\Lambda$\textsf{(}$\Phi^{1}$\textsf{,MAHC,HC)}, but in this case $\Phi^{1}=$ \textsf{5Br(Hu,MAHC,CEM)} (which will be denoted from now on as \textsf{Ca}, for reasons of simplicity) is the best meta-cooperative algorithm found in level $1$. Table \ref{tab:resulCa5} shows the computational results for the $16$ instances of the ToSP for this depth of meta-cooperation. In general, all \textsf{5}$\Lambda$\textsf{(Ca,$\star$,$\star$)} combinations (where $\star$ represents any basic algorithm) outperforms the results of previous meta-cooperative models. 
\end{sloppypar}

\begin{sloppypar}
As in the previous scenario, and in order to analyze the significance of the results, but also to obtain a global perspective on how compare with each other the algorithms, we use a rank-based approach as previously explained. The distribution of these ranks are shown in Figure \ref{fig:Fi1Fi2}. It can be seen that $\Phi^{2}$ models outperform $\Phi^{1}$ models, in particular \textsf{5}$\Lambda$\textsf{(Ca,MAHC,CEM)} and \textsf{5}$\Lambda$\textsf{(Ca,CEM,HC)}. But also, as in previous scenario, using CEM at 0-level implies better performance over the other basic algorithm combinations that do no use it.
This is also confirmed by the statistical tests. Firstly, the output of Quade test for this scenario shows that the statistic values (both for all algorithms, and for the top-ten) are clearly higher than the critical values ($p$-value $\approx 0$ again), hence indicating the global differences are statistically significant. Then, Holm test is conducted using \textsf{5Br(Ca,MAHC,CEM)} --the algorithm with best mean rank-- as control algorithm. As shown in Table \ref{statistical:holm2}, a statistically significant difference is noted with respect to variants that do not use \textsf{CEM} or that use the \textsf{Random} topology (and conversely, there is no statistically significant difference with variants using \textsf{CEM} and \textsf{Broadcast}/\textsf{Ring} topology). This underpins the importance of building on top components from lower layers of the architectural hierarchy, as well as the relevance of a more regular communication topology (observe in this regard how \textsf{Random} was not profusely represented in the top-10 for 1-level models --Table \ref{statistical:holm}-- and such models were found to be outperformed by the control algorithm with statistical significance). 
\end{sloppypar}%

\begin{figure}[!t]
\includegraphics[width=\textwidth]{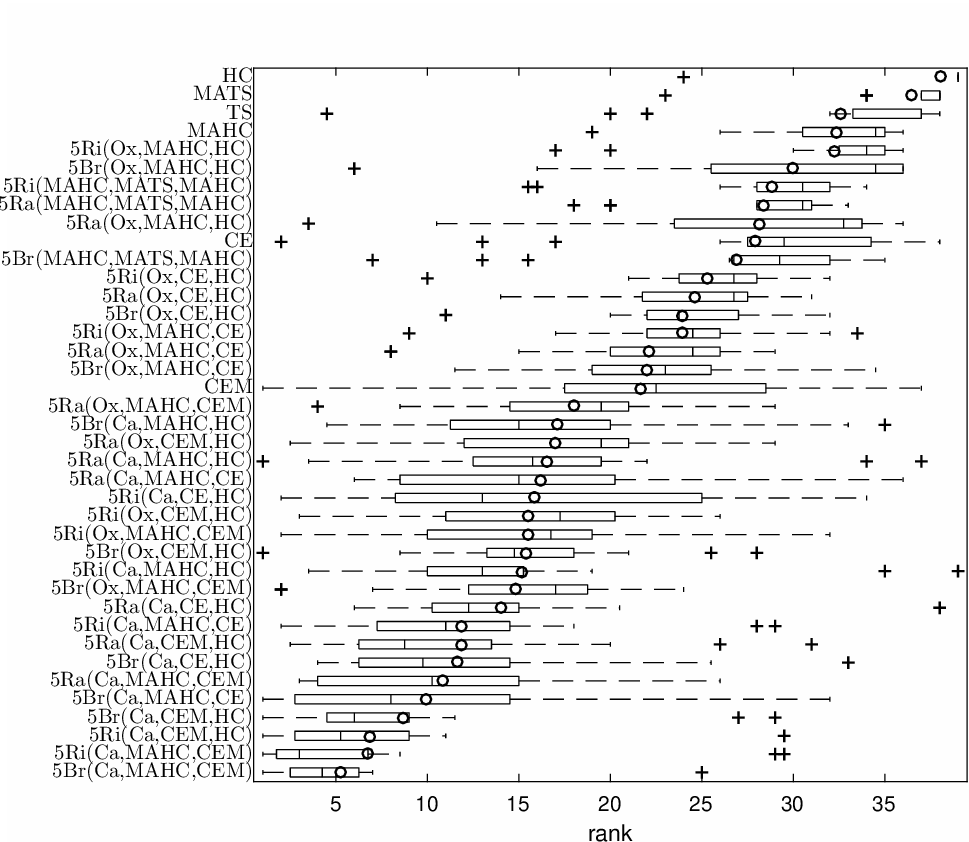}
\caption{Rank distribution for 3-level-meta-cooperative models. %
}
\label{fig:Fi2Fi3}
\end{figure}

\subsubsection{Scenario with 3-level-meta-cooperative models }

\begin{sloppypar}
Results from previous section shown that the control algorithm uses \textsf{Ca}, \textsf{MAHC} and \textsf{CEM} with \textsf{Broadcast} topology. We have thus used this algorithm as building block to construct the next level. As in previous sections, we use for simplicity the denomination \textsf{Ox} for this algorithm $\Phi^{2}=$ \textsf{5Br(Ca,MAHC,CEM)}. Thus, we have used \textsf{Ox} as $\Phi^{2}$ to execute the five schemes described above, i.e., \textsf{5}$\Lambda$\textsf{($\Phi^{2}$,CE,HC)}, \textsf{5}$\Lambda$\textsf{($\Phi^{2}$,CEM,HC)}, \textsf{5}$\Lambda$\textsf{($\Phi^{2}$,MAHC,CE)}, \textsf{5}$\Lambda$\textsf{($\Phi^{2}$,MAHC,CEM)} and \textsf{5}$\Lambda$\textsf{($\Phi^{2}$,MAHC,HC)}. The computational results are shown in Table \ref{tab:resulOx5}. A first observation is immediately obtained: the performance of $\Phi^{3}$ models is worse than that of algorithms based on the 2-level meta-cooperation scheme \textsf{5Br(Ca,$\star$,$\star$)}. This is also reflected in the rank-based analysis depicted in Figure \ref{fig:Fi2Fi3}. No $\Phi^{3}$ model makes it into the top-10 section (the best \textsf{Ox}-based variant ranks in 13th position).
 Indeed, Quade test is passed both for the whole collection of algorithms and for the top-ten algorithms ($p$-value $\approx 0$), indicating in this case that $\Phi^{3}$ models are significantly outperformed by $\Phi^{2}$ algorithms\footnote{It must be noted that Holm test would in this case be equivalent to the test performed in the previous section since the top-ten algorithms are exactly the same ones.}. The model seems to have hit the maximum depth level at which useful cooperation is established. Clearly, the increased depth has a toll on the computational budget allocated to each component of the particular instance considered. As a result, these cannot take advantage of such a small budget. Basically, the search effort is overdiversified in light of the computational challenge they had to face. 

\end{sloppypar}

\subsubsection{Comparison of topologies used in the meta-cooperative models }

\begin{figure}[!t]
\subfloat[]{\includegraphics[width=.5\textwidth]{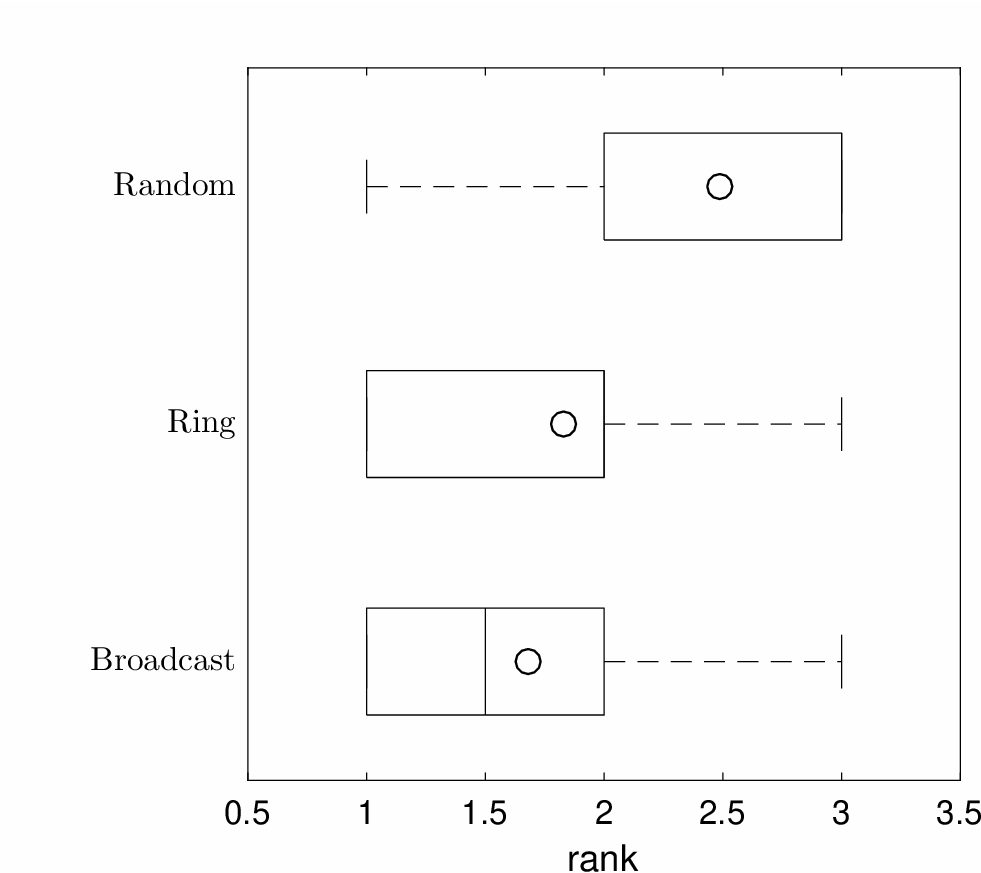}}
\subfloat[]{\includegraphics[width=.5\textwidth]{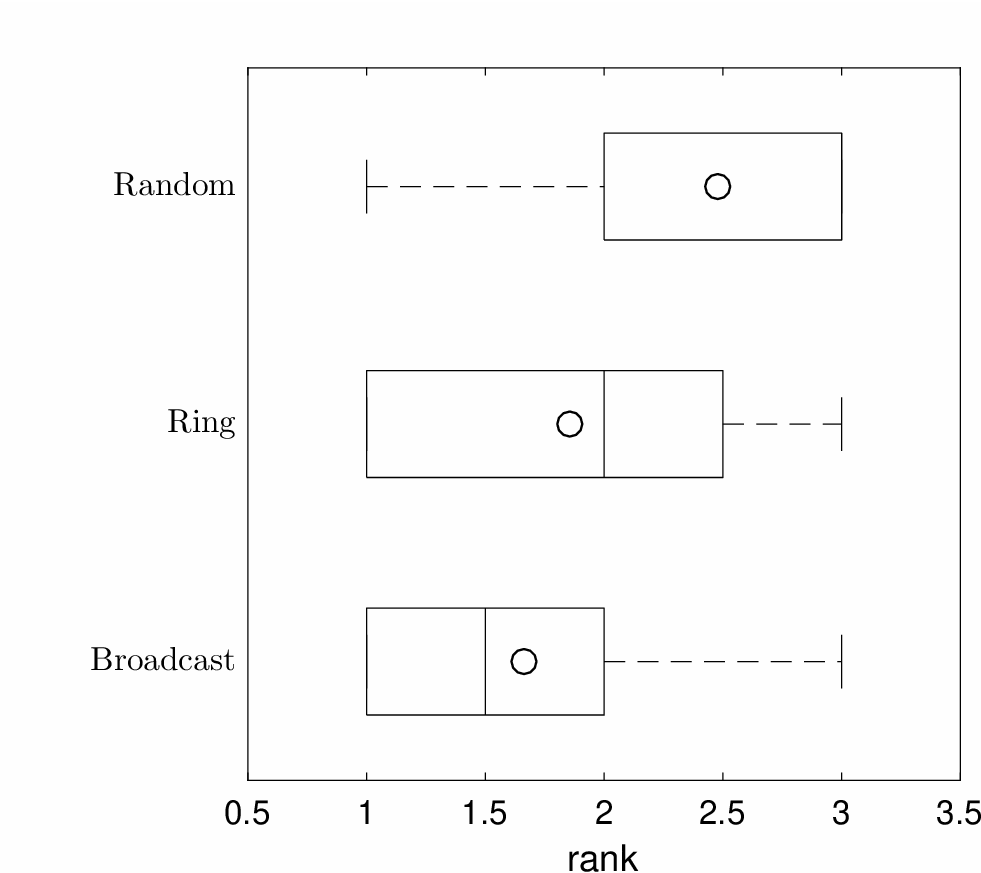}}

\subfloat[]{\includegraphics[width=.5\textwidth]{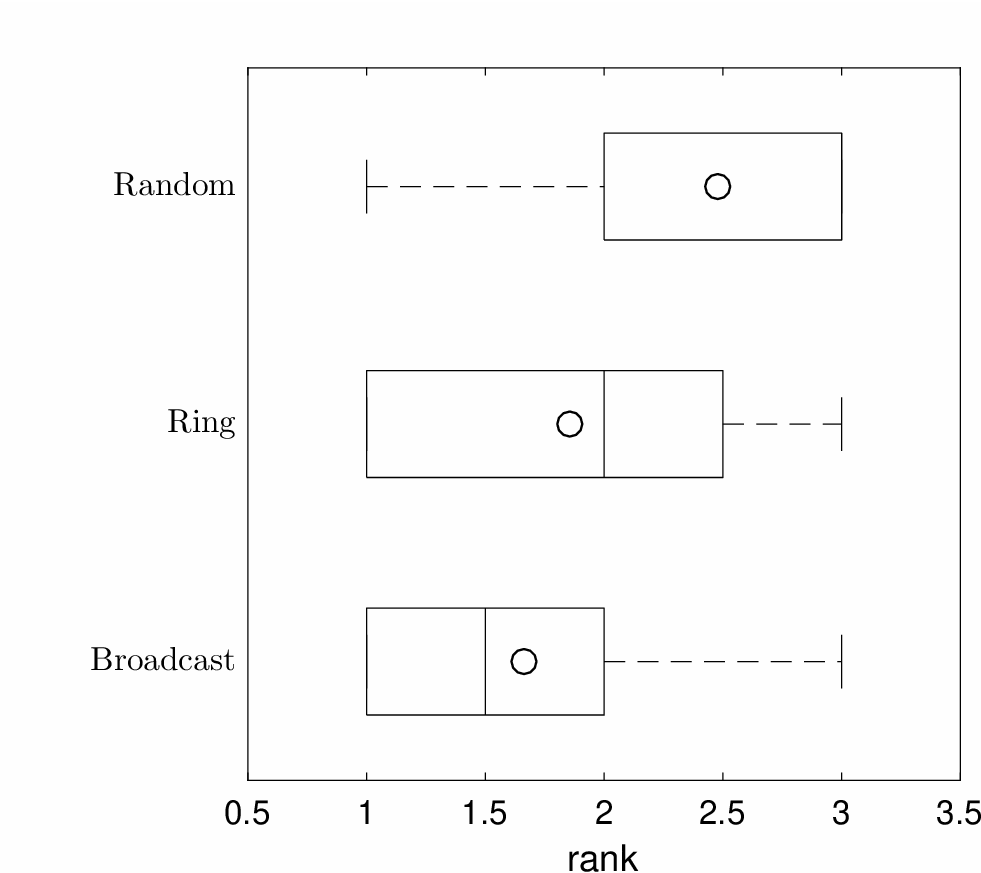}}
\subfloat[]{\includegraphics[width=.5\textwidth]{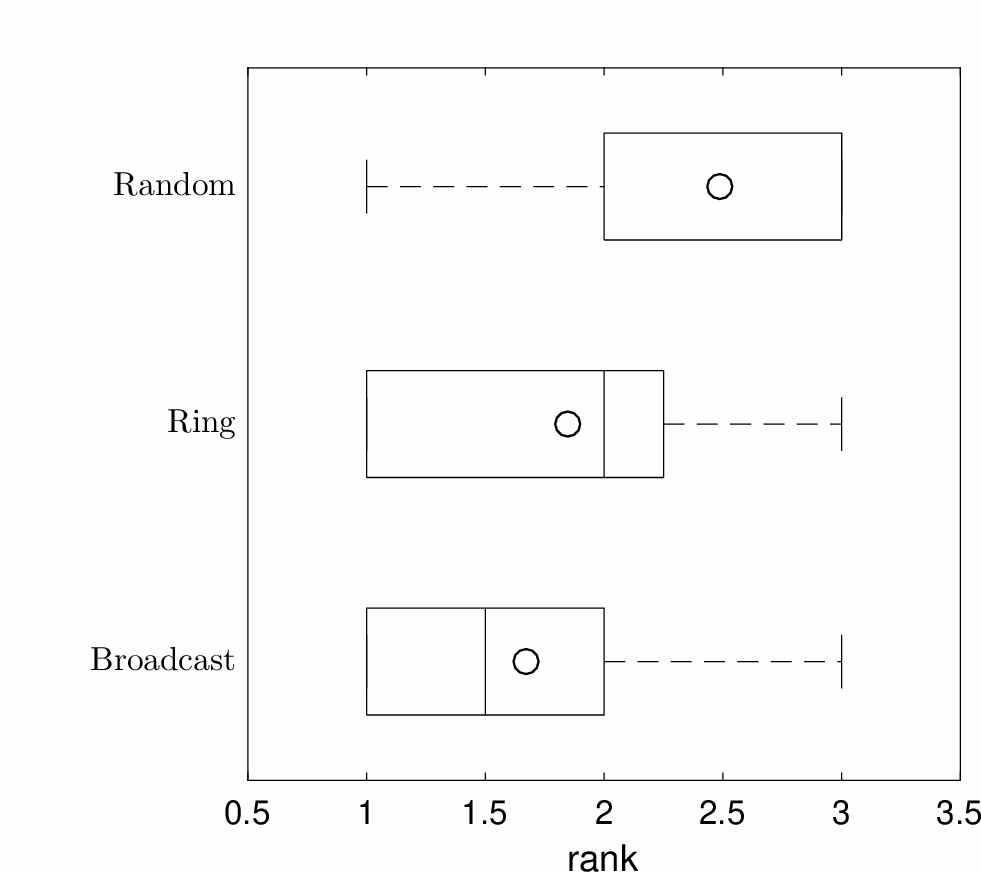}}
\caption{Rank distribution of each interconnection topology. (a) 1-level (b) 2-level (c) 3-level (d) global for all models
}
\label{fig:CoopComparative}
\end{figure}

\begin{sloppypar}
In order to determine the topology that best adapts to the meta-cooperative
models considered, we proceed to perform an analysis of the results obtained, factorizing along this topological dimension for the highest level of cooperation. To this end, we use the same rank-based analysis utilized in previous section, comparing in this case the three topologies for each combination of constituent algorithms and problem instance. This analysis has been performed individually per scenario (1-level, 2-level and 3-level), as well as globally for all of them. 
\end{sloppypar}

\begin{sloppypar}
Figure \ref{fig:CoopComparative} shows the outcome of the analysis. The relative ranking is consistently \textsf{Broadcast}$<$\textsf{Ring}$<$\textsf{Random}. Indeed, in all cases Quade test is passed with $p$-value $\approx0$. When the scenarios are analyzed separately, Holm test indicates that \textsf{Broadcast} (the control topology with the best average performance) is significantly better than \textsf{Random} (see Tables \ref{tab:topo1}--\ref{tab:topo3}). As anticipated before, the irregularity of this latter topology tends to reduce the intensification of the algorithm as a whole, as opposed to the more exploitative \textsf{Broadcast} topology. The \textsf{Ring} topology lies in the middle ground, not being significantly worse than \textsf{Broadcast} in each of the three scenarios independently ($p$-values between $0.11$ and $0.17$). However, when the data for all scenarios is aggregated, the superiority of \textsf{Broadcast} can be determined over \textsf{Ring} as well ($p$-value $\approx0.025$) as shown in Table \ref{tab:topoglobal}. This underpins the importance of establishing strong synchronization within the involved techniques as an intensification factor to balance the inherent diversification provided by the diversity of search algorithms at multiple levels. 
\end{sloppypar}

{\color{black}
\subsubsection{Discussion}

 There are two major high-level factors affecting the balance between intensification and diversification in the model: the number of agents and the depth of meta-cooperation. As indicated before, for a given fixed computational budget, having a higher number of agents implies each of them receives a lower amount of resources for conducting the search. At some point, such resources may not be enough for attaining adequate convergence to high-quality regions of the search space (or at least, not enough for establishing an adequate cooperation and benefiting from migrated information). This is also important in light of the depth of the model, since the latter will typically correlate with a higher number of agents (hence the previous issue would be applicable). Notice also that deeper architectures will also result in a more complex spatial structure through which information diffusion will be more gradual. This also contributes to a more diversified search. The designer has in this sense several degrees of freedom to try to find an adequate balance. Firstly, picking agents whose search profile provide higher intensification (as it the case of e.g., \textsf{CEM} in this context) in order to counterbalance to some extent the larger diversification. Secondly, adjusting the interconnection topology for faster diffusion of information (hence, the better performance of the \textsf{Broadcast} topology from a global perspective here). Of course, increasing the intensification at any level within the system can eventually result in premature convergence of individual agents, so one can expect performance to hit a plateau and decrease onward upon reaching a certain depth (which will depend on the nature of the problem considered and the techniques involved).}

\section{Conclusions \label{sec:conclusions}}

Deep meta-cooperative optimization models are an appropriate 
model
for the integration of different search techniques, each of them
with a different view of the search space. Thus combining
different search patterns and levels of meta-cooperation, the process can take advantage from an increased
ability to escape from local optima and focus on promising regions
of the search space. %
{\color{black} As in deep learning models, interpretability is an issue. Being structurally-complex models, design decisions can have non-linear effects in performance. While this might highlight the need for further studies on the dynamics of the search, it also points towards the incorporation of self-$\star$ properties \citep{Babaoglu2005,Berns2009} as a means for self-managing the optimization process. Also, it is clear that this kind of models may only be useful in scenarios where individual search algorithms are not powerful enough on their own, so as to avoid using a sledgehammer to crack nuts, wasting computational resources in unproductive explorations.}

\begin{sloppypar}
An empirical evaluation performed on the ToSP, a hard combinatorial optimization
problem, provides evidence of the 
validity and efficiency of the proposed techniques. The results show that 
for this problem an architecture with two nested levels of meta-cooperation
(i.e., with a depth of three algorithmic layers)
provides
better computational results than well known algorithms for the resolution
of the problem, such as basic memetic algorithms, cross-entropy methods and 
single-level cooperative methods. Moreover, the results show that it is possible to generate
schemes that improve the overall performance of these methods in particular
by including them in a deeper meta-cooperative scheme. Indeed,
the experimentation indicates that 2-level-meta-cooperative
algorithms outperform the algorithms of levels $0$ (state-of-the art 
cooperative algorithms applied to handle the ToSP) and $1$. Performance
degrades when going beyond this point, indicating that the depth of meta-cooperation needs to be tuned depending on factors such as the scale of the problem and possibly the structure of the search landscape as well.
\end{sloppypar}

Future work will extend the experimental setup by putting the model to test under different scales, be it regarding the problem instances or the model itself. A forensic analysis of the population statistics will be conducted to determine the search behavior of each component of the hierarchy. {\color{black}The selection of migrants is another factor whose influence should be studied, since the use of different strategies (random selection, selection based on diversity, etc.) may play an important role in adjusting the intensification/diversification balance of the algorithm.} Parallelization of the method is also of the foremost interest: as the number of evaluations in each cycle is divided by the number of agents, different execution times are expected, and therefore, different asynchronous communication mechanisms and topologies, implying comparisons and load balancing experiments, would be required. {\color{black} This is naturally amenable to the incorporation of self-adaptive capabilities whereby parameters or even the system structure itself is subject to change according to what has been found by the algorithms up-to-now. Results from hierarchical genetic search systems \citep{DBLP:conf/foga/SchaeferK02,Schaefer12hierarchical} can pave the way to this end.}

\section*{Acknowledgements}
{\color{black}The authors wish to thank the anonymous reviewers for their helpful comments.} The first author thanks to the Decanato de Investigaci\'{o}n of UNET the partial support of the present research. Second and third author were partially supported by Universidad de M\'{a}laga, Campus de Excelencia Internacional Andaluc\'{i}a Tech, and also by 
research projects Ephemech\footnote{\url{https://ephemech.wordpress.com/}} (TIN2014-56494-C4-1-P), and DeepBio\footnote{\url{https://deepbio.wordpress.com}} (TIN2017-85727-C4-01-P), funded by Ministerio Espa\~{n}ol de Econom\'{\i}a y Competitividad. Fourth author was also partially supported by ``Ayuda del Programa de Fomento e Impulso de la Actividad Investigadora de la Universidad de C\'adiz''.

\appendix

\section{Tables of Computational Results}\label{App:tables}
This appendix section shows the computational results for each level in Tables \ref{tab:resulHu5}, \ref{tab:resulCa5}, and \ref{tab:resulOx5} respectively.

\begin{landscape}
\begin{table}[!ht]
\caption{Computational results with HC, TS, memetic algorithms with HC ---as proposed in \citep{Amaya2008}---, cooperative models $cycles_{max}\Lambda(MAHC,MATS,MAHC)$---as proposed in \citep{Amaya2011MC}--- standard CE and CEM with 4 PMFS. $\bar x$ = mean number of tool switches. $\sigma$ = mean standard deviation.}
\label{tab:resulBef}
\begin{scriptsize}
\begin{center}
\begin{tabular}{ r  r  r  r  r  r  r  r  r  r  r  r  r  r  r  r  r  r  r  r  r  r  r  r}
\hline
    &       &   \begin{sideways}\textsf{HC}\end{sideways}    &    \begin{sideways}\textsf{TS}\end{sideways}   &    \begin{sideways}\textsf{MAHC}\end{sideways}    &   \begin{sideways}\textsf{MATS}\end{sideways}   &    \begin{sideways}\textsf{5Br(MAHC,MATS,MAHC)}\end{sideways}    &   \begin{sideways}\textsf{5Ra(MAHC,MATS,MAHC)}\end{sideways} &    \begin{sideways}\textsf{5Ri(MAHC,MATS,MAHC)}\end{sideways} &    \begin{sideways}\textsf{CE}\end{sideways} &    \begin{sideways}\textsf{CEM}\end{sideways}      \\
\hline\hline
$4\zeta_{10}^{9}$	&	$\bar x$	&	8.4	&	8.08	&	8.1	&	8.14	&	8.04	&	 8.0	 &	7.98	&	8.12	&	8.06	\\		
	&	$\sigma$	&	1.0	&	0.74	&	0.75	&	0.87	&	0.82	&	0.82	 &	0.79	&	0.36	&	0.44	\\		
\hline
$4\zeta_{10}^{10}$	&	$\bar x$	&	9.6	&	8.8	&	8.94	&	9.08	&	8.74	&	 8.8	 &	8.86	&	9.04	&	9.04	\\		
	&	$\sigma$	&	1.57	&	1.61	&	1.62	&	1.66	&	1.61	&	 1.6	 &	1.71	&	 0.43	&	0.26	\\		
\hline
$6\zeta_{10}^{15}$	&	$\bar x$	&	14.7	&	13.68	&	13.89	&	13.86	&	 13.72	 &	13.82	&	13.76	&	14.02	&	14.0	\\		
	&	$\sigma$	&	2.25	&	2.1	&	1.99	&	2.03	&	2.05	&	1.97	 &	2.11	&	0.47	&	0.51	\\		
\hline
$6\zeta_{15}^{12}$	&	$\bar x$	&	20.1	&	16.46	&	16.26	&	16.9	&	 15.94	 &	16.04	&	16.12	&	16.04	&	15.64	\\		
	&	$\sigma$	&	2.05	&	1.93	&	1.79	&	2.07	&	1.96	&	 2.03	 &	1.82	&	0.9	&	0.63	\\		
\hline
$6\zeta_{15}^{20}$	&	$\bar x$	&	26.54	&	23.02	&	23.18	&	23.14	&	 22.78	 &	22.92	&	22.84	&	23.3	&	23.26	\\		
	&	$\sigma$	&	2.4	&	2.0	&	1.96	&	2.01	&	1.91	&	1.85	 &	 2.04	&	1.0	&	0.9	\\		
\hline
$8\zeta_{20}^{15}$	&	$\bar x$	&	28.9	&	23.62	&	22.86	&	24.46	&	23.0	 &	22.92	&	22.9	&	22.28	&	21.58	\\		
	&	$\sigma$	&	4.17	&	3.63	&	3.41	&	3.62	&	3.44	&	 3.52	 &	3.37	&	0.89	&	0.7	\\		
\hline
$8\zeta_{20}^{16}$	&	$\bar x$	&	33.78	&	27.92	&	27.24	&	28.66	&	 27.12	 &	26.74	&	26.78	&	26.18	&	25.78	\\		
	&	$\sigma$	&	2.48	&	2.13	&	2.22	&	2.08	&	2.02	&	 2.13	 &	1.96	&	1.09	&	0.67	\\		
\hline
$10\zeta_{20}^{20}$	&	$\bar x$	&	37.46	&	30.72	&	30.53	&	31.38	&	 30.16	 &	30.16	&	30.26	&	30.28	&	29.28	\\		
	&	$\sigma$	&	2.88	&	2.5	&	2.49	&	2.33	&	2.38	&	2.34	 &	2.38	&	0.69	&	0.72	\\		
\hline
$10\zeta_{30}^{25}$	&	$\bar x$	&	85.46	&	67.72	&	64.32	&	68.24	&	 64.58	 &	64.92	&	64.3	&	60.24	&	59.04	\\		
	&	$\sigma$	&	2.97	&	1.52	&	2.4	&	1.84	&	1.89	&	1.98	 &	1.93	&	1.46	&	0.97	\\		
\hline
$15\zeta_{30}^{40}$	&	$\bar x$	&	121.2	&	101.72	&	99.7	&	102.34	&	99.2	 &	99.14	&	98.64	&	96.72	&	94.94	\\		
	&	$\sigma$	&	14.61	&	13.07	&	12.82	&	12.46	&	12.8	&	 12.99	&	12.41	&	1.59	&	1.28	\\		
\hline
$15\zeta_{40}^{30}$	&	$\bar x$	&	127.54	&	101.9	&	95.86	&	103.3	&	95.1	 &	95.26	&	95.46	&	88.62	&	86.3	\\		
	&	$\sigma$	&	9.51	&	8.14	&	7.52	&	8.26	&	6.95	&	 7.93	 &	7.62	&	1.33	&	1.32	\\		
\hline
$20\zeta_{40}^{60}$	&	$\bar x$	&	248.7	&	213.74	&	206.3	&	212.2	&	 206.2	 &	205.66	&	206.0	&	199.1	&	195.98	\\		
	&	$\sigma$	&	10.17	&	8.38	&	8.81	&	8.73	&	8.41	&	 8.27	 &	7.92	&	2.13	&	2.01	\\		
\hline
$24\zeta_{20}^{30}$	&	$\bar x$	&	30.24	&	25.04	&	24.78	&	25.76	&	 24.34	 &	24.66	&	24.34	&	23.98	&	23.46	\\		
	&	$\sigma$	&	3.69	&	3.02	&	3.29	&	3.55	&	3.05	&	 3.5	 &	3.1	&	0.83	&	0.59	\\		
\hline
$24\zeta_{20}^{36}$	&	$\bar x$	&	53.24	&	45.9	&	44.87	&	46.71	&	45.0	 &	44.98	&	44.92	&	45.12	&	43.96	\\		
	&	$\sigma$	&	9.33	&	8.98	&	7.55	&	9.4	&	8.1	&	7.98	 &	 8.02	&	1.28	&	0.98	\\		
\hline
$25\zeta_{50}^{40}$	&	$\bar x$	&	191.02	&	153.58	&	144.18	&	157.44	&	 143.38	 &	143.92	&	144.72	&	131.88	&	130.76	\\		
	&	$\sigma$	&	13.36	&	12.89	&	11.94	&	12.07	&	12.31	&	 12.76	&	11.67	&	2.0	&	1.74	\\		
\hline
$30\zeta_{20}^{40}$	&	$\bar x$	&	49.12	&	42.12	&	41.3	&	42.06	&	 40.86	 &	40.8	&	41.04	&	40.84	&	40.08	\\		
	&	$\sigma$	&	5.29	&	4.34	&	4.41	&	5.17	&	4.25	&	 4.4	 &	4.66	&	1.24	&	1.1	\\		
\hline
\end{tabular}
\end{center}
\end{scriptsize}
\end{table}
\end{landscape}


\begin{landscape}
\begin{table}[!ht]
\caption{Computational results with meta-cooperative models \textsf{5}$\Lambda$\textsf{(Hu, ..., ...)} where \textsf{Hu=5Ri(MAHC,MATS,MAHC)}. $\bar x$ = mean number of tool switches. $\sigma$ = mean standard deviation. Recall we are using the notation $C\zeta_{m}^{n}$, where C is the magazine capacity, m is the total number of tools and n is the number of jobs. }
\label{tab:resulHu5}
\begin{scriptsize}
\begin{center}
\begin{tabular}{ r  r  r  r  r  r  r  r  r  r  r  r  r  r  r  r  r  r  r  r  r  r  r  r}
\hline
    &       &   \begin{sideways}\textsf{5Br(Hu,CE,HC)}\end{sideways}    &    \begin{sideways}\textsf{5Br(Hu,CEM,HC)}\end{sideways}   &    \begin{sideways}\textsf{5Br(Hu,MAHC,CE)}\end{sideways}    &   \begin{sideways}\textsf{5Br(Hu,MAHC,CEM)}\end{sideways}   &    \begin{sideways}\textsf{5Br(Hu,MAHC,HC)}\end{sideways}   &    \begin{sideways}\textsf{5Ra(Hu,CE,HC)}\end{sideways}   &    \begin{sideways}\textsf{5Ra(Hu,CEM,HC)}\end{sideways}    &   \begin{sideways}\textsf{5Ra(Hu,MAHC,CE)}\end{sideways}    &   \begin{sideways}\textsf{5Ra(Hu,MAHC,CEM)}\end{sideways} &    \begin{sideways}\textsf{5Ra(Hu,MAHC,HC)}\end{sideways} &    \begin{sideways}\textsf{5Ri(Hu,CE,HC)}\end{sideways} &    \begin{sideways}\textsf{5Ri(Hu,CEM,HC)}\end{sideways} &    \begin{sideways}\textsf{5Ri(Hu,MAHC,CE)}\end{sideways} &    \begin{sideways}\textsf{5Ri(Hu,MAHC,CEM)}\end{sideways} &    \begin{sideways}\textsf{5Ri(Hu,MAHC,HC)}\end{sideways}   \\
\hline\hline
$4\zeta_{10}^{9}$	&	$\bar x$	&	8.06	&	7.96	&	8.02	&	8.04	&	8.0	 &	 8.0	&	7.96	&	8.02	&	8.08	&	8.0	&	8.08	&	7.98	&	 8.1	&	 8.08	 &	7.98	\\
	&	$\sigma$	&	0.79	&	0.85	&	0.73	&	0.77	&	0.77	&	 0.72	 &	0.72	&	0.79	&	0.82	&	0.87	&	0.8	&	0.76	&	0.75	 &	0.84	 &	0.73	\\
\hline
$4\zeta_{10}^{10}$	&	$\bar x$	&	8.8	&	8.9	&	8.88	&	8.82	&	8.88	&	 8.86	 &	8.74	&	8.9	&	8.88	&	8.86	&	8.9	&	8.72	&	8.88	 &	 8.8	&	 8.9	\\
	&	$\sigma$	&	1.67	&	1.68	&	1.54	&	1.69	&	1.76	&	 1.66	 &	1.69	&	1.64	&	1.58	&	1.76	&	1.7	&	1.65	&	1.61	 &	1.59	 &	1.73	\\
\hline
$6\zeta_{10}^{15}$	&	$\bar x$	&	13.86	&	13.72	&	13.76	&	13.68	&	 13.92	 &	13.74	&	13.76	&	13.72	&	13.74	&	13.72	&	13.78	&	 13.72	&	 13.72	&	13.7	&	13.84	\\
	&	$\sigma$	&	1.98	&	2.15	&	2.05	&	2.16	&	2.05	&	 2.07	 &	2.07	&	2.13	&	2.06	&	2.11	&	2.12	&	2.06	&	 2.12	&	 2.12	&	2.06	\\
\hline
$6\zeta_{15}^{12}$	&	$\bar x$	&	15.6	&	15.38	&	15.56	&	15.32	&	 16.44	 &	15.9	&	15.44	&	15.7	&	15.44	&	16.14	&	15.78	&	 15.44	&	 15.46	&	15.3	&	16.18	\\
	&	$\sigma$	&	1.92	&	1.92	&	1.8	&	1.86	&	1.99	&	1.88	 &	1.79	&	1.96	&	1.69	&	1.94	&	1.93	&	1.95	&	1.87	 &	 1.84	&	1.79	\\
\hline
$6\zeta_{15}^{20}$	&	$\bar x$	&	22.66	&	22.58	&	22.72	&	22.44	&	 23.08	 &	22.64	&	22.62	&	22.7	&	22.28	&	22.9	&	22.76	&	 22.34	&	 22.54	&	22.5	&	23.06	\\
	&	$\sigma$	&	1.87	&	1.86	&	1.84	&	1.89	&	2.07	&	 1.82	 &	1.71	&	1.98	&	1.81	&	2.08	&	1.96	&	1.78	&	 1.95	&	 1.96	&	2.2	\\
\hline
$8\zeta_{20}^{15}$	&	$\bar x$	&	21.84	&	21.36	&	21.56	&	21.16	&	 22.84	 &	21.92	&	21.58	&	21.88	&	21.44	&	23.12	&	21.8	&	 21.12	&	 21.9	&	21.3	&	23.12	\\
	&	$\sigma$	&	3.32	&	3.33	&	3.29	&	3.28	&	3.56	&	 3.2	 &	3.25	&	3.27	&	3.35	&	3.48	&	3.29	&	3.21	&	3.27	 &	 3.32	&	3.56	\\
\hline
$8\zeta_{20}^{16}$	&	$\bar x$	&	25.86	&	25.34	&	25.86	&	25.38	&	 27.06	 &	26.28	&	25.48	&	25.98	&	25.38	&	27.0	&	25.94	&	 25.56	&	 25.88	&	25.38	&	27.0	\\
	&	$\sigma$	&	1.97	&	1.88	&	1.83	&	1.74	&	2.11	&	 1.59	 &	1.7	&	1.95	&	1.77	&	2.16	&	2.03	&	1.59	&	1.98	 &	1.68	 &	2.14	\\
\hline
$10\zeta_{20}^{20}$	&	$\bar x$	&	29.32	&	28.74	&	29.32	&	28.74	&	 30.64	 &	29.5	&	29.06	&	29.5	&	28.84	&	30.5	&	29.28	&	 28.82	&	 29.42	&	28.74	&	30.32	\\
	&	$\sigma$	&	2.29	&	2.16	&	2.41	&	2.27	&	2.3	&	2.41	 &	2.23	&	2.21	&	2.19	&	2.39	&	2.27	&	2.35	&	2.21	 &	 2.41	&	2.59	\\
\hline
$10\zeta_{30}^{25}$	&	$\bar x$	&	59.76	&	58.86	&	60.08	&	58.86	&	 65.14	 &	60.28	&	59.1	&	60.06	&	59.4	&	64.82	&	60.2	&	 59.04	&	 60.08	&	59.04	&	64.76	\\
	&	$\sigma$	&	1.19	&	1.04	&	1.26	&	1.02	&	2.21	&	 1.37	 &	0.92	&	1.59	&	1.06	&	2.23	&	1.33	&	1.04	&	 1.55	&	 1.11	&	2.85	\\
\hline
$15\zeta_{30}^{40}$	&	$\bar x$	&	95.44	&	93.2	&	95.0	&	93.58	&	 99.06	 &	95.62	&	93.68	&	95.34	&	93.64	&	99.02	&	95.04	&	 93.6	&	 95.42	&	93.1	&	98.52	\\
	&	$\sigma$	&	12.4	&	12.45	&	12.52	&	12.28	&	12.75	&	 12.31	&	12.33	&	12.3	&	12.4	&	12.61	&	12.49	&	12.43	&	 13.21	&	12.33	&	12.29	\\
\hline
$15\zeta_{40}^{30}$	&	$\bar x$	&	90.38	&	91.2	&	90.52	&	90.62	&	 97.38	 &	91.02	&	91.62	&	90.58	&	91.5	&	96.72	&	90.12	&	 91.58	&	 90.24	&	90.84	&	95.58	\\
	&	$\sigma$	&	7.89	&	7.75	&	7.59	&	7.78	&	8.11	&	 7.96	 &	7.53	&	7.66	&	7.42	&	7.7	&	6.79	&	7.38	&	7.71	 &	7.66	 &	7.54	\\
\hline
$20\zeta_{40}^{60}$	&	$\bar x$	&	196.72	&	194.76	&	197.1	&	194.54	&	 206.58	 &	197.98	&	195.44	&	197.8	&	195.5	&	205.56	&	197.42	&	 195.02	&	 197.42	&	194.86	&	205.46	\\
	&	$\sigma$	&	7.62	&	7.76	&	8.41	&	7.59	&	8.21	&	 7.51	 &	7.11	&	8.2	&	7.42	&	8.47	&	8.23	&	7.8	&	7.65	 &	 7.15	 &	8.32	\\
\hline
$24\zeta_{20}^{30}$	&	$\bar x$	&	33.73	&	33.16	&	33.54	&	33.14	&	 34.99	 &	33.92	&	33.26	&	33.76	&	33.25	&	35.23	&	34.0	&	 33.17	&	 33.63	&	33.01	&	35.08	\\
	&	$\sigma$	&	11.74	&	11.62	&	11.74	&	11.56	&	12.27	&	 11.71	&	11.6	&	11.65	&	11.49	&	12.16	&	11.89	&	11.43	&	 11.58	&	11.49	&	12.11	\\
\hline
$24\zeta_{20}^{36}$	&	$\bar x$	&	43.62	&	42.9	&	43.64	&	42.64	&	 45.18	 &	44.0	&	43.38	&	43.82	&	43.34	&	45.2	&	43.72	&	 43.26	&	 43.72	&	42.96	&	45.04	\\
	&	$\sigma$	&	8.07	&	7.58	&	8.1	&	7.69	&	8.63	&	8.17	 &	7.24	&	7.67	&	7.82	&	8.75	&	7.72	&	7.59	&	7.79	 &	 7.7	&	8.4	\\
\hline
$25\zeta_{50}^{40}$	&	$\bar x$	&	143.04	&	139.96	&	142.3	&	139.14	&	 144.74	 &	144.28	&	140.68	&	144.1	&	140.18	&	145.12	&	143.6	&	 139.88	&	 142.36	&	139.2	&	145.22	\\
	&	$\sigma$	&	12.06	&	11.52	&	10.98	&	10.99	&	12.06	&	 11.58	&	10.6	&	11.75	&	11.36	&	12.72	&	11.1	&	11.04	&	 11.76	&	11.74	&	11.98	\\
\hline
$30\zeta_{20}^{40}$	&	$\bar x$	&	40.34	&	39.3	&	39.6	&	39.14	&	 41.52	 &	40.36	&	39.52	&	40.14	&	39.5	&	41.36	&	39.98	&	 39.24	&	 40.12	&	39.34	&	41.08	\\
	&	$\sigma$	&	4.45	&	4.48	&	4.56	&	4.28	&	4.66	&	 4.48	 &	4.49	&	4.39	&	4.5	&	4.6	&	4.44	&	4.62	&	4.52	 &	 4.21	 &	4.37	\\
\hline
\end{tabular}
\end{center}
\end{scriptsize}
\end{table}
\end{landscape}


\begin{landscape}
\begin{table}[!ht]
\caption{Computational results with meta-cooperative models \textsf{5}$\Lambda$\textsf{(Ca, $\star$, $\star$)} where \textsf{Ca=Br(Hu,MAHC,CEM)}. $\bar x$ = mean number of tool switches. $\sigma$ = mean standard deviation. Recall we are using the notation $C\zeta_{m}^{n}$, where C is the magazine capacity, m is the total number of tools and n is the number of jobs. }
\label{tab:resulCa5}
\begin{scriptsize}
\begin{center}
\begin{tabular}{ r  r  r  r  r  r  r  r  r  r  r  r  r  r  r  r  r  r  r  r  r  r  r  r}
\hline
    &       &   \begin{sideways}\textsf{5Br(Ca,CE,HC)}\end{sideways}    &    \begin{sideways}\textsf{5Br(Ca,CEM,HC)}\end{sideways}   &    \begin{sideways}\textsf{5Br(Ca,MAHC,CE)}\end{sideways}    &   \begin{sideways}\textsf{5Br(Ca,MAHC,CEM)}\end{sideways}   &    \begin{sideways}\textsf{5Br(Ca,MAHC,HC)}\end{sideways}   &    \begin{sideways}\textsf{5Ra(Ca,CE,HC)}\end{sideways}   &    \begin{sideways}\textsf{5Ra(Ca,CEM,HC)}\end{sideways}    &   \begin{sideways}\textsf{5Ra(Ca,MAHC,CE)}\end{sideways}    &   \begin{sideways}\textsf{5Ra(Ca,MAHC,CEM)}\end{sideways} &    \begin{sideways}\textsf{5Ra(Ca,MAHC,HC)}\end{sideways} &    \begin{sideways}\textsf{5Ri(Ca,CE,HC)}\end{sideways} &    \begin{sideways}\textsf{5Ri(Ca,CEM,HC)}\end{sideways} &    \begin{sideways}\textsf{5Ri(Ca,MAHC,CE)}\end{sideways} &    \begin{sideways}\textsf{5Ri(Ca,MAHC,CEM)}\end{sideways} &    \begin{sideways}\textsf{5Ri(Ca,MAHC,HC)}\end{sideways}   \\
\hline\hline
$4\zeta_{10}^{9}$	&	$\bar x$	&	7.94	&	7.92	&	8.0	&	7.92	&	8.0	&	 7.92	 &	7.9	&	8.02	&	8.02	&	7.98	&	8.02	&	7.92	&	8.04	 &	7.88	 &	7.96	\\
	&	$\sigma$	&	0.81	&	0.8	&	0.77	&	0.8	&	0.77	&	0.72	 &	 0.81	&	0.73	&	0.84	&	0.79	&	0.73	&	0.77	&	0.72	 &	 0.77	 &	0.82	\\
\hline
$4\zeta_{10}^{10}$	&	$\bar x$	&	8.84	&	8.76	&	8.82	&	8.68	&	8.76	 &	8.74	&	8.8	&	8.9	&	8.76	&	8.7	&	8.86	&	8.7	&	8.78	&	 8.86	 &	8.7	\\
	&	$\sigma$	&	1.71	&	1.67	&	1.55	&	1.63	&	1.68	&	 1.69	 &	1.78	&	1.63	&	1.56	&	1.63	&	1.67	&	1.68	&	 1.57	&	 1.61	&	1.7	\\
\hline
$6\zeta_{10}^{15}$	&	$\bar x$	&	13.68	&	13.78	&	13.72	&	13.68	&	 13.82	 &	13.74	&	13.76	&	13.74	&	13.74	&	13.64	&	13.78	&	 13.68	&	 13.72	&	13.7	&	13.72	\\
	&	$\sigma$	&	2.11	&	2.03	&	2.05	&	2.09	&	2.06	&	 2.11	 &	2.11	&	2.06	&	2.06	&	2.13	&	2.13	&	2.11	&	 2.1	 &	 2.13	 &	2.08	\\
\hline
$6\zeta_{15}^{12}$	&	$\bar x$	&	15.34	&	15.02	&	15.22	&	15.12	&	 15.22	 &	15.16	&	15.08	&	15.14	&	15.04	&	15.32	&	15.04	&	 15.04	&	 15.22	&	15.0	&	15.28	\\
	&	$\sigma$	&	1.76	&	1.97	&	1.76	&	1.84	&	1.77	&	 1.88	 &	1.79	&	1.71	&	1.75	&	1.88	&	1.87	&	1.73	&	 1.8	 &	 1.84	 &	1.84	\\
\hline
$6\zeta_{15}^{20}$	&	$\bar x$	&	22.24	&	21.96	&	22.06	&	22.12	&	 22.44	 &	22.32	&	22.2	&	22.14	&	22.08	&	22.44	&	22.32	&	 22.18	&	 22.14	&	22.06	&	22.22	\\
	&	$\sigma$	&	1.69	&	1.74	&	1.8	&	1.76	&	1.76	&	1.71	 &	1.71	&	1.73	&	1.79	&	1.7	&	1.86	&	1.82	&	1.72	&	 1.87	 &	1.74	\\
\hline
$8\zeta_{20}^{15}$	&	$\bar x$	&	20.94	&	20.96	&	20.68	&	20.8	&	 21.02	 &	21.02	&	20.82	&	20.9	&	20.8	&	21.28	&	21.14	&	 20.8	&	 21.0	&	20.82	&	21.02	\\
	&	$\sigma$	&	3.24	&	3.22	&	3.19	&	3.34	&	3.37	&	 3.33	 &	3.31	&	3.27	&	3.28	&	3.09	&	3.12	&	3.3	&	3.28	 &	3.14	 &	3.33	\\
\hline
$8\zeta_{20}^{16}$	&	$\bar x$	&	24.94	&	24.94	&	24.96	&	24.86	&	 25.32	 &	25.08	&	24.96	&	25.04	&	25.04	&	25.24	&	24.8	&	 24.78	&	 24.96	&	24.86	&	25.08	\\
	&	$\sigma$	&	1.78	&	1.67	&	1.6	&	1.64	&	1.78	&	1.65	 &	1.65	&	1.77	&	1.57	&	1.86	&	1.56	&	1.64	&	1.64	 &	 1.52	&	1.8	\\
\hline
$10\zeta_{20}^{20}$	&	$\bar x$	&	28.42	&	28.44	&	28.38	&	28.48	&	 28.72	 &	28.52	&	28.52	&	28.64	&	28.6	&	28.68	&	28.6	&	 28.46	&	 28.34	&	28.32	&	28.56	\\
	&	$\sigma$	&	2.01	&	2.17	&	2.35	&	2.06	&	2.31	&	 2.3	 &	2.36	&	2.2	&	2.32	&	2.12	&	2.19	&	2.14	&	2.29	 &	 2.27	 &	2.27	\\
\hline
$10\zeta_{30}^{25}$	&	$\bar x$	&	58.5	&	58.32	&	58.08	&	58.14	&	 58.66	 &	58.62	&	58.36	&	58.4	&	58.18	&	59.62	&	58.3	&	 58.14	&	 58.56	&	58.18	&	58.64	\\
	&	$\sigma$	&	0.94	&	0.61	&	0.89	&	0.85	&	0.99	&	 1.07	 &	0.74	&	1.1	&	0.93	&	2.46	&	0.85	&	0.92	&	1.12	 &	0.77	 &	0.62	\\
\hline
$15\zeta_{30}^{40}$	&	$\bar x$	&	93.1	&	92.92	&	93.0	&	92.96	&	 93.42	 &	93.46	&	92.96	&	93.62	&	92.94	&	93.48	&	93.1	&	 92.66	&	 92.74	&	92.78	&	93.58	\\
	&	$\sigma$	&	12.51	&	12.25	&	12.3	&	12.37	&	12.42	&	 12.05	&	12.07	&	12.36	&	12.38	&	12.37	&	12.47	&	12.28	&	 12.52	&	12.42	&	12.83	\\
\hline
$15\zeta_{40}^{30}$	&	$\bar x$	&	87.0	&	86.98	&	86.8	&	86.92	&	 86.96	 &	87.44	&	87.36	&	87.26	&	87.52	&	87.74	&	87.06	&	 87.14	&	 86.74	&	87.06	&	87.08	\\
	&	$\sigma$	&	7.5	&	7.22	&	7.79	&	7.16	&	7.04	&	7.7	 &	 7.24	&	7.76	&	7.85	&	7.85	&	7.16	&	7.02	&	7.19	 &	 6.85	 &	7.37	\\
\hline
$20\zeta_{40}^{60}$	&	$\bar x$	&	195.8	&	194.84	&	195.02	&	194.36	&	 196.12	 &	195.66	&	194.84	&	195.86	&	194.46	&	196.62	&	196.28	&	 194.36	&	 195.44	&	194.34	&	196.06	\\
	&	$\sigma$	&	7.47	&	7.36	&	7.24	&	7.41	&	7.93	&	 7.61	 &	7.65	&	7.5	&	7.6	&	7.86	&	7.32	&	7.3	&	7.62	&	 7.39	&	 7.43	\\
\hline
$24\zeta_{20}^{30}$	&	$\bar x$	&	32.74	&	32.55	&	32.72	&	32.5	&	 32.78	 &	33.04	&	32.65	&	32.84	&	32.53	&	32.9	&	32.77	&	 32.62	&	 32.58	&	32.62	&	33.09	\\
	&	$\sigma$	&	11.58	&	11.35	&	11.5	&	11.35	&	11.44	&	 11.6	 &	11.3	&	11.55	&	11.42	&	11.67	&	11.49	&	11.44	&	 11.34	&	 11.44	&	11.69	\\
\hline
$24\zeta_{20}^{36}$	&	$\bar x$	&	43.68	&	42.98	&	43.72	&	42.72	&	 45.22	 &	43.9	&	43.32	&	43.94	&	43.2	&	45.44	&	44.2	&	 43.02	&	 43.88	&	42.84	&	45.46	\\
	&	$\sigma$	&	7.9	&	7.64	&	7.97	&	7.89	&	8.23	&	8.12	 &	7.29	&	7.51	&	7.61	&	8.47	&	7.76	&	7.5	&	7.96	&	 7.56	 &	8.47	\\
\hline
$25\zeta_{50}^{40}$	&	$\bar x$	&	133.5	&	133.66	&	133.42	&	133.38	&	 133.38	 &	135.38	&	134.24	&	135.5	&	134.5	&	133.98	&	134.22	&	 133.72	&	 133.88	&	133.34	&	133.56	\\
	&	$\sigma$	&	11.26	&	10.76	&	10.35	&	11.25	&	11.53	&	 11.2	 &	10.83	&	11.53	&	10.78	&	10.91	&	10.67	&	11.03	&	 10.78	&	 10.25	&	10.71	\\
\hline
$30\zeta_{20}^{40}$	&	$\bar x$	&	38.88	&	38.96	&	38.76	&	38.5	&	 39.18	 &	39.1	&	39.12	&	38.94	&	38.82	&	39.1	&	39.0	&	 39.06	&	 39.02	&	38.78	&	39.14	\\
	&	$\sigma$	&	4.34	&	4.45	&	4.47	&	4.29	&	4.51	&	 4.38	 &	4.51	&	4.63	&	4.31	&	4.3	&	4.43	&	4.28	&	4.32	 &	4.33	 &	4.46	\\
\hline
\end{tabular}
\end{center}
\end{scriptsize}
\end{table}
\end{landscape}


\begin{landscape}
\begin{table}[!ht]
\caption{Computational results with meta-cooperative models \textsf{5}$\Lambda$\textsf{(Ox, $\star$, $\star$)} where \textsf{Ox=5Br(Ca,MAHC,CEM)}. $\bar x$ = mean number of tool switches. $\sigma$ = mean standard deviation. Recall we are using the notation $C\zeta_{m}^{n}$, where C is the magazine capacity, m is the total number of tools and n is the number of jobs. }
\label{tab:resulOx5}
\begin{scriptsize}
\begin{center}
\begin{tabular}{ r  r  r  r  r  r  r  r  r  r  r  r  r  r  r  r  r  r  r  r  r  r  r  r}
\hline
    &       &   \begin{sideways}\textsf{5Br(Ox,CE,HC)}\end{sideways}    &    \begin{sideways}\textsf{5Br(Ox,CEM,HC)}\end{sideways}   &    \begin{sideways}\textsf{5Br(Ox,MAHC,CE)}\end{sideways}    &   \begin{sideways}\textsf{5Br(Ox,MAHC,CEM)}\end{sideways}   &    \begin{sideways}\textsf{5Br(Ox,MAHC,HC)}\end{sideways}   &    \begin{sideways}\textsf{5Ra(Ox,CE,HC)}\end{sideways}   &    \begin{sideways}\textsf{5Ra(Ox,CEM,HC)}\end{sideways}    &   \begin{sideways}\textsf{5Ra(Ox,MAHC,CE)}\end{sideways}    &   \begin{sideways}\textsf{5Ra(Ox,MAHC,CEM)}\end{sideways} &    \begin{sideways}\textsf{5Ra(Ox,MAHC,HC)}\end{sideways} &    \begin{sideways}\textsf{5Ri(Ox,CE,HC)}\end{sideways} &    \begin{sideways}\textsf{5Ri(Ox,CEM,HC)}\end{sideways} &    \begin{sideways}\textsf{5Ri(Ox,MAHC,CE)}\end{sideways} &    \begin{sideways}\textsf{5Ri(Ox,MAHC,CEM)}\end{sideways} &    \begin{sideways}\textsf{5Ri(Ox,MAHC,HC)}\end{sideways}   \\
\hline\hline
$4\zeta_{10}^{9}$	&	$\bar x$	&	8.04	&	7.96	&	8.08	&	8.0	&	7.92	 &	 8.0	&	7.9	&	8.02	&	7.98	&	7.94	&	8.02	&	7.94	&	 8.06	&	 7.94	&	8.06	\\
	&	$\sigma$	&	0.77	&	0.75	&	0.82	&	0.8	&	0.77	&	0.82	 &	0.75	&	0.73	&	0.73	&	0.73	&	0.76	&	0.79	&	0.81	 &	 0.79	&	0.81	\\
\hline
$4\zeta_{10}^{10}$	&	$\bar x$	&	8.8	&	8.84	&	8.76	&	8.76	&	8.78	 &	 8.78	&	8.74	&	8.82	&	8.86	&	8.7	&	8.86	&	8.76	&	 8.9	&	 8.88	&	8.8	\\
	&	$\sigma$	&	1.66	&	1.76	&	1.63	&	1.66	&	1.63	&	 1.69	 &	1.71	&	1.68	&	1.55	&	1.69	&	1.72	&	1.66	&	 1.65	&	 1.56	&	1.71	\\
\hline
$6\zeta_{10}^{15}$	&	$\bar x$	&	13.82	&	13.72	&	13.78	&	13.74	&	 13.74	 &	13.74	&	13.72	&	13.7	&	13.72	&	13.74	&	13.76	&	 13.7	&	 13.74	&	13.66	&	13.9	\\
	&	$\sigma$	&	2.1	&	2.07	&	2.02	&	2.06	&	2.12	&	2.07	 &	2.09	&	2.06	&	2.09	&	2.1	&	2.11	&	2.08	&	2.12	&	 2.09	 &	2.09	\\
\hline
$6\zeta_{15}^{12}$	&	$\bar x$	&	15.56	&	15.28	&	15.68	&	15.4	&	 16.18	 &	15.76	&	15.54	&	15.6	&	15.4	&	16.04	&	15.72	&	 15.44	&	 15.56	&	15.32	&	16.1	\\
	&	$\sigma$	&	1.98	&	1.88	&	1.76	&	1.88	&	1.99	&	 1.78	 &	1.9	&	1.87	&	1.83	&	2.08	&	1.95	&	1.68	&	1.69	 &	1.77	 &	1.89	\\
\hline
$6\zeta_{15}^{20}$	&	$\bar x$	&	22.7	&	22.34	&	22.52	&	22.32	&	 23.22	 &	22.78	&	22.72	&	22.54	&	22.48	&	22.78	&	22.78	&	 22.6	&	 22.82	&	22.44	&	23.06	\\
	&	$\sigma$	&	1.79	&	1.73	&	1.69	&	1.88	&	2.01	&	 1.77	 &	1.77	&	1.7	&	1.86	&	1.94	&	1.84	&	1.92	&	1.89	 &	1.83	 &	1.89	\\
\hline
$8\zeta_{20}^{15}$	&	$\bar x$	&	21.72	&	21.36	&	21.78	&	21.4	&	 23.12	 &	21.94	&	21.52	&	21.96	&	21.46	&	23.0	&	21.72	&	 21.08	&	 21.76	&	21.32	&	23.1	\\
	&	$\sigma$	&	3.47	&	3.39	&	3.66	&	3.35	&	3.75	&	 3.25	 &	3.28	&	3.19	&	3.16	&	3.76	&	3.32	&	3.18	&	 3.37	&	 3.3	&	3.53	\\
\hline
$8\zeta_{20}^{16}$	&	$\bar x$	&	26.0	&	25.24	&	25.72	&	25.38	&	 27.32	 &	26.1	&	25.56	&	25.88	&	25.62	&	27.16	&	26.08	&	 25.26	&	 26.02	&	25.52	&	27.18	\\
	&	$\sigma$	&	1.79	&	1.62	&	1.73	&	1.51	&	2.41	&	 2.06	 &	1.8	&	1.72	&	1.73	&	2.08	&	1.92	&	1.71	&	1.46	 &	1.71	 &	2.0	\\
\hline
$10\zeta_{20}^{20}$	&	$\bar x$	&	29.38	&	28.8	&	29.4	&	28.78	&	 30.68	 &	29.68	&	29.02	&	29.48	&	29.04	&	30.5	&	29.52	&	 29.04	&	 29.12	&	28.74	&	30.54	\\
	&	$\sigma$	&	2.42	&	2.25	&	2.34	&	2.36	&	2.36	&	 2.62	 &	2.19	&	2.23	&	2.31	&	2.44	&	2.28	&	2.24	&	 2.43	&	 2.3	&	2.37	\\
\hline
$10\zeta_{30}^{25}$	&	$\bar x$	&	60.3	&	59.02	&	60.08	&	58.72	&	 65.32	 &	60.28	&	59.08	&	60.44	&	59.3	&	65.58	&	60.12	&	 59.08	&	 60.2	&	59.16	&	65.56	\\
	&	$\sigma$	&	1.3	&	0.81	&	1.28	&	1.13	&	1.97	&	1.23	 &	1.06	&	1.22	&	1.06	&	2.34	&	1.37	&	1.0	&	1.17	&	 1.07	 &	2.3	\\
\hline
$15\zeta_{30}^{40}$	&	$\bar x$	&	95.4	&	93.54	&	95.0	&	93.54	&	 99.92	 &	95.84	&	94.22	&	95.92	&	93.64	&	99.42	&	96.08	&	 93.7	&	 95.36	&	93.58	&	99.5	\\
	&	$\sigma$	&	12.26	&	12.32	&	12.61	&	12.67	&	13.16	&	 12.52	&	11.82	&	12.43	&	12.21	&	12.46	&	12.44	&	12.46	&	 12.28	&	12.29	&	12.73	\\
\hline
$15\zeta_{40}^{30}$	&	$\bar x$	&	90.44	&	91.92	&	90.12	&	91.38	&	96.5	 &	90.86	&	92.02	&	90.4	&	91.68	&	96.72	&	90.58	&	91.48	 &	 90.6	&	91.4	&	97.8	\\
	&	$\sigma$	&	7.0	&	7.38	&	7.72	&	7.47	&	7.38	&	7.86	 &	7.82	&	7.39	&	7.87	&	7.88	&	7.49	&	7.75	&	7.29	 &	 7.28	&	7.78	\\
\hline
$20\zeta_{40}^{60}$	&	$\bar x$	&	197.9	&	195.2	&	196.84	&	195.0	&	 205.86	 &	198.82	&	195.82	&	197.7	&	195.26	&	205.86	&	198.86	&	 195.42	&	 197.62	&	195.2	&	205.34	\\
	&	$\sigma$	&	7.6	&	7.79	&	7.73	&	7.63	&	8.08	&	7.55	 &	7.74	&	7.87	&	7.6	&	8.25	&	7.63	&	7.61	&	7.94	&	 7.87	 &	8.44	\\
\hline
$24\zeta_{20}^{30}$	&	$\bar x$	&	23.74	&	23.04	&	23.78	&	23.16	&	 24.96	 &	24.0	&	23.28	&	23.48	&	23.24	&	24.86	&	23.7	&	 23.2	&	 23.52	&	23.26	&	24.62	\\
	&	$\sigma$	&	3.18	&	3.0	&	3.11	&	3.17	&	3.35	&	3.05	 &	3.08	&	3.29	&	3.13	&	3.26	&	2.93	&	3.17	&	3.28	 &	 3.08	&	3.43	\\
\hline
$24\zeta_{20}^{36}$	&	$\bar x$	&	43.96	&	43.14	&	43.32	&	42.8	&	 45.08	 &	44.14	&	43.3	&	43.7	&	43.14	&	45.14	&	44.0	&	 43.0	&	 43.86	&	43.0	&	45.68	\\
	&	$\sigma$	&	7.82	&	7.79	&	7.95	&	7.75	&	8.02	&	 8.11	 &	7.65	&	8.02	&	7.73	&	8.34	&	7.53	&	7.74	&	 7.81	&	 7.77	&	8.46	\\
\hline
$25\zeta_{50}^{40}$	&	$\bar x$	&	144.28	&	140.38	&	142.26	&	138.82	&	 146.06	 &	144.52	&	141.1	&	142.64	&	140.14	&	145.1	&	144.64	&	 140.5	&	 142.7	&	139.08	&	145.6	\\
	&	$\sigma$	&	10.96	&	11.12	&	12.0	&	11.29	&	11.78	&	 11.74	&	11.62	&	12.32	&	10.47	&	12.54	&	10.55	&	11.65	&	 11.21	&	11.38	&	12.3	\\
\hline
$30\zeta_{20}^{40}$	&	$\bar x$	&	39.98	&	39.24	&	39.86	&	39.28	&	41.5	 &	40.32	&	39.56	&	40.26	&	39.6	&	41.52	&	40.48	&	39.52	 &	 40.12	&	39.32	&	41.42	\\
	&	$\sigma$	&	4.53	&	4.3	&	4.38	&	4.51	&	4.27	&	4.27	 &	4.45	&	4.41	&	4.27	&	4.52	&	4.61	&	4.55	&	4.58	 &	 4.51	&	4.17	\\

\hline
\end{tabular}
\end{center}
\end{scriptsize}
\end{table}
\end{landscape}

\section{Tests}
\label{app:tests}
This appendix shows the results of Holm test for each level in Tables \ref{statistical:holm} and \ref{statistical:holm2} (Table for level 3 is equivalent to the table for level 2, as they share the best ten algorithms in the comparison). Holm test for the topology in each of the three scenarios and globally are also shown in Tables \ref{tab:topo1}--\ref{tab:topoglobal}.

\begin{table}[!hb]
\caption{Results of Holm test ($\alpha=0.05$) using \textsf{5Br(Hu,MAHC,CEM)} as control algorithm. }
\label{statistical:holm}
\centering{}%
\begin{tabular}{llrrrr}
$i$ & strategy         & ~\hspace{5mm}~$z$-statistic & ~\hspace{5mm}~$p$-value & ~\hspace{9mm}~$\alpha/i$ & \\\hline
1 & \textsf{5Ri(Hu,MAHC,CEM)} & 7.298e-01 & 2.327e-01 & 5.000e-02 & fail \\ 
2 & \textsf{5Br(Hu,CEM,HC)} & 9.634e-01 & 1.677e-01 & 2.500e-02 & fail \\ 
3 & \textsf{5Ri(Hu,CEM,HC)} & 1.255e+00 & 1.047e-01 & 1.667e-02 & fail \\ 
\hline
4 & \textsf{5Ra(Hu,MAHC,CEM)} & 2.803e+00 & 2.535e-03 & 1.250e-02 & \\ 
5 & \textsf{5Ra(Hu,CEM,HC)} & 3.153e+00 & 8.082e-04 & 1.000e-02 & \\ 
6 & \textsf{CEM} & 3.883e+00 & 5.164e-05 & 8.333e-03 & \\ 
7 & \textsf{5Br(Hu,MAHC,CE)} & 4.642e+00 & 1.727e-06 & 7.143e-03 & \\ 
8 & \textsf{5Br(Hu,CE,HC)} & 5.109e+00 & 1.620e-07 & 6.250e-03 & \\ 
9 & \textsf{5Ri(Hu,MAHC,CE)} & 5.196e+00 & 1.015e-07 & 5.556e-03 & \\ 
\hline
\end{tabular}
\end{table}

\begin{table}[!hb]
\caption{Results of Holms test ($\alpha=0.05$) using \textsf{5Br(Ca,MAHC,CEM)} as
control algorithm.}
\label{statistical:holm2}
\centering{}
\begin{tabular}{llrrrr}
$i$ & strategy         & ~\hspace{5mm}~$z$-statistic & ~\hspace{5mm}~$p$-value & ~\hspace{9mm}~$\alpha/i$ & \\\hline
1 & \textsf{5Ri(Ca,MAHC,CEM)} & 1.752e-01 & 4.305e-01 & 5.000e-02 & fail \\ 
2 & \textsf{5Ri(Ca,CEM,HC)} & 8.758e-01 & 1.906e-01 & 2.500e-02 & fail \\ 
3 & \textsf{5Br(Ca,CEM,HC)} & 1.489e+00 & 6.826e-02 & 1.667e-02 & fail \\ 
4 & \textsf{5Br(Ca,MAHC,CE)} & 1.868e+00 & 3.085e-02 & 1.250e-02 & fail \\ 
\hline
5 & \textsf{5Ra(Ca,MAHC,CEM)} & 2.540e+00 & 5.545e-03 & 1.000e-02 & \\ 
6 & \textsf{5Ri(Ca,MAHC,CE)} & 2.949e+00 & 1.596e-03 & 8.333e-03 & \\ 
7 & \textsf{5Br(Ca,CE,HC)} & 3.445e+00 & 2.857e-04 & 7.143e-03 & \\ 
8 & \textsf{5Ra(Ca,CEM,HC)} & 3.474e+00 & 2.563e-04 & 6.250e-03 & \\ 
9 & \textsf{5Ra(Ca,CE,HC)} & 5.080e+00 & 1.890e-07 & 5.556e-03 & \\ 
\hline
\end{tabular}
\end{table}

\begin{table}[!hb]
\caption{Results of Holm Test ($\alpha=0.05$) using \textsf{Broadcast} as control algorithm on the 1-level scenario. \label{tab:topo1}}
\begin{tabular}{llrrrr}
$i$ & strategy         & ~\hspace{5mm}~$z$-statistic & ~\hspace{5mm}~$p$-value & ~\hspace{9mm}~$\alpha/i$ & \\\hline
1 & \textsf{Ring} & 9.487e-01 & 1.714e-01 & 5.000e-02 & fail \\ 
\hline
2 & \textsf{Random} & 5.099e+00 & 1.706e-07 & 2.500e-02 & \\ 
\hline
\end{tabular}
\end{table}

\begin{table}[!hb]
\caption{Results of Holm Test ($\alpha=0.05$) using \textsf{Broadcast} as control algorithm on the 2-level scenario. \label{tab:topo2}}
\begin{tabular}{llrrrr}
$i$ & strategy         & ~\hspace{5mm}~$z$-statistic & ~\hspace{5mm}~$p$-value & ~\hspace{9mm}~$\alpha/i$ & \\\hline
1 & \textsf{Ring} & 1.225e+00 & 1.102e-01 & 5.000e-02 & fail \\ 
\hline
2 & \textsf{Random} & 5.178e+00 & 1.120e-07 & 2.500e-02 & \\ 
\hline
\end{tabular}
\end{table}

\begin{table}[!hb]
\caption{Results of Holm Test ($\alpha=0.05$) using \textsf{Broadcast} as control algorithm on the 3-level scenario. \label{tab:topo3}}
\begin{tabular}{llrrrr}
$i$ & strategy         & ~\hspace{5mm}~$z$-statistic & ~\hspace{5mm}~$p$-value & ~\hspace{9mm}~$\alpha/i$ & \\\hline
1 & \textsf{Ring} & 1.225e+00 & 1.102e-01 & 5.000e-02 & fail \\ 
\hline
2 & \textsf{Random} & 5.178e+00 & 1.120e-07 & 2.500e-02 & \\ 
\hline
\end{tabular}
\end{table}

\begin{table}[!hb]
\caption{Results of Holm Test ($\alpha=0.05$) using \textsf{Broadcast} as control algorithm across all three scenarios. \label{tab:topoglobal}}
\begin{tabular}{llrrrr}
$i$ & strategy         & ~\hspace{5mm}~$z$-statistic & ~\hspace{5mm}~$p$-value & ~\hspace{9mm}~$\alpha/i$ & \\\hline
1 & \textsf{Ring} & 1.963e+00 & 2.484e-02 & 5.000e-02 & \\ 
2 & \textsf{Random} & 8.923e+00 & 2.263e-19 & 2.500e-02 & \\ 
\hline
\end{tabular}
\end{table}

\end{document}